\documentclass{article}

\usepackage[table]{xcolor} 
\usepackage{microtype}
\usepackage{graphicx}
\usepackage{booktabs} 

\usepackage{hyperref}

\usepackage[accepted]{icml2024}

\usepackage{amsmath}
\usepackage{amssymb}
\usepackage{mathtools}
\usepackage{amsthm}

\usepackage[capitalize,noabbrev]{cleveref}

\theoremstyle{plain}

\theoremstyle{definition}

\theoremstyle{remark}

\usepackage[textsize=tiny]{todonotes}

\icmltitlerunning{Autoencoding Conditional Neural Processes for Representation Learning}

\usepackage{tikz} 
\usepackage{float}       
\usepackage{subcaption}
\usepackage{bm}
\usepackage{caption}
\usepackage{multirow}
\usepackage{paralist}
\usepackage{wrapfig}
\usepackage[most]{tcolorbox}
\usepackage{xurl}
\usepackage{circledsteps}
\usepackage{pstricks}

\usetikzlibrary{calc,backgrounds,positioning,bayesnet,graphs,arrows,fit,matrix}
\usetikzlibrary{shapes.geometric,patterns,decorations.pathreplacing}
\def\figuresize{8mm}

\tikzset{ob/.style={obs,minimum size=\figuresize}}
\tikzset{det/.style={latent,rectangle,rounded corners=1pt,minimum size=\figuresize}}
\tikzset{lat/.style={latent,minimum size=\figuresize}}
\tikzset{bgb/.style={rounded corners=1mm,minimum height=8mm,minimum width=1.2cm}}
\tikzset{|-/.style={to path={|- (\tikztotarget)}}}
\tikzset{-|/.style={to path={-| (\tikztotarget)}}}
\tikzset{>={stealth},shorten >=1pt}

\newcommand{\ppsvae}{\textsc{PPS-\!VAE}}
\newcommand{\ppsvaea}{\ppsvae\textsuperscript{a}}
\newcommand{\ppsvaei}{\ppsvae\textsuperscript{i}}
\newcommand{\E}{\mathbb{E}}
\newcommand{\KL}[2]{D_{\text{KL}}\left(#1 \Vert #2\right)}
\newcommand{\X}{\mathcal{X}}
\newcommand{\Y}{\mathcal{Y}}
\newcommand{\C}{\mathcal{C}}
\newcommand{\N}{\mathcal{N}}
\newcommand{\xm}{\bm{x}_M}
\newcommand{\ym}{\bm{y}_M}
\newcommand{\xt}{\bm{x}_T}
\newcommand{\yt}{\bm{y}_T}

\makeatletter
\renewcommand\paragraph{\@startsection{paragraph}{4}{\z@}%
{0.2ex \@plus.1ex \@minus.1ex}%
{-1em}%
{\normalfont\normalsize\bfseries}}
\makeatother

\setdefaultleftmargin{2ex}{2ex}{}{}{}{}
\begin{document}
\twocolumn[
\icmltitle{Autoencoding Conditional Neural Processes for Representation Learning}



\icmlsetsymbol{equal}{*}

\begin{icmlauthorlist}
\icmlauthor{Victor Prokhorov}{yyy}
\icmlauthor{Ivan Titov}{yyy,sch}
\icmlauthor{N. Siddharth}{yyy,comp}
\end{icmlauthorlist}

\icmlaffiliation{yyy}{School of Informatics, University of Edinburgh}
\icmlaffiliation{comp}{The Alan Turing Institute}
\icmlaffiliation{sch}{ILLC, University of Amsterdam}

\icmlcorrespondingauthor{Victor Prokhorov}{victorprokhorov91@gmail.com}


\vskip 0.3in
]



\printAffiliationsAndNotice{}  



\begin{abstract}
Conditional neural processes (CNPs) are a flexible and efficient family of models that \emph{learn to learn} a stochastic process from data. 
They have seen particular application in contextual image completion---observing pixel values at some locations to predict a distribution over values at other unobserved locations. 
However, the choice of pixels in learning CNPs is typically either random or derived from a simple statistical measure (e.g. pixel variance). 
Here, we turn the problem on its head and ask: which pixels would a CNP like to observe---do
they facilitate fitting better CNPs, and do such pixels tell us something meaningful about the underlying image?
To this end we develop the Partial Pixel Space Variational Autoencoder (\ppsvae), an amortised variational framework that casts CNP context as latent variables learnt simultaneously with the CNP.
We evaluate \ppsvae\ over a number of tasks across different visual data, and find that not only can it facilitate better-fit CNPs, but also that the spatial arrangement and values meaningfully characterise image information---evaluated through the lens of classification on both within and out-of-data distributions.
Our model additionally allows for \emph{dynamic} adaption of context-set size and the ability to scale-up to larger images, providing a promising avenue to explore learning meaningful and effective visual representations.

\end{abstract}

\vspace*{-2\baselineskip}
\section{Introduction}
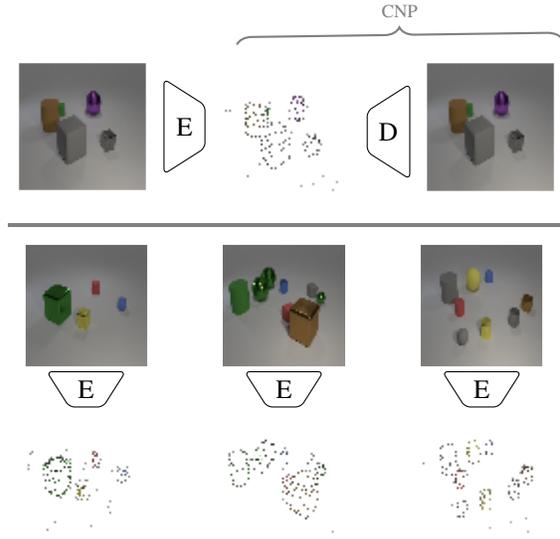
\begin{figure}[t!]{}
  \centering
  \begin{tikzpicture}[trap/.style={trapezium,trapezium angle=57.5,draw,minimum width=8ex,rounded
    corners=2pt}]
    \node[inner sep=0] (ip)
      {\includegraphics[width=0.23\linewidth]{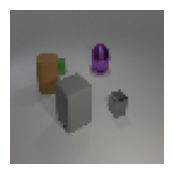}};
    \node[trap,right=4mm of ip,rotate=-90,anchor=center] (e) {\rotatebox{90}{E}};
    \node[inner sep=0, right=4mm of e,yshift=4mm] (pps)
      {\includegraphics[width=0.23\linewidth]{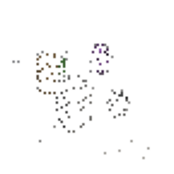}};
    \node[trap,right=4mm of pps,rotate=90,anchor=center] (d) {\rotatebox{-90}{D}};
    \node[inner sep=0, right=4mm of d,yshift=-4mm] (op)
      {\includegraphics[width=0.23\linewidth]{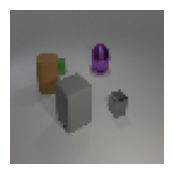}};
    \draw [thick, decoration={brace,raise=2mm,amplitude=2mm}, decorate, gray]
      ($(pps.north west)+(3mm,0)$) -- (op.north east)
      node[pos=0.5,above=4mm]{\scriptsize CNP};
  \end{tikzpicture}
  \textcolor{gray}{\rule[1mm]{0.9\linewidth}{1.5pt}}
  \begin{tikzpicture}[trap/.style={trapezium,trapezium angle=57.5,draw,minimum width=6ex,rounded
    corners=2pt}]
    \node[inner sep=0] (ip1)
      {\includegraphics[width=0.22\linewidth]{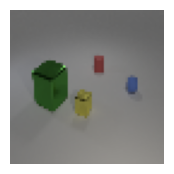}};
    \node[trap,below=2mm of ip1,rotate=180,anchor=center] (e1) {\rotatebox{180}{E}};
    \node[inner sep=0, below=5mm of e1]
      {\includegraphics[width=0.22\linewidth]{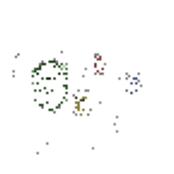}};    
    \node[inner sep=0, right=8mm of ip1] (ip2)
      {\includegraphics[width=0.22\linewidth]{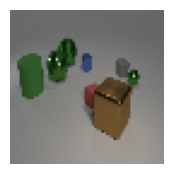}};
    \node[trap,below=2mm of ip2,rotate=180,anchor=center] (e2) {\rotatebox{180}{E}};
    \node[inner sep=0, below=5mm of e2]
      {\includegraphics[width=0.22\linewidth]{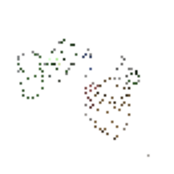}};    
    \node[inner sep=0, right=8mm of ip2] (ip3)
      {\includegraphics[width=0.22\linewidth]{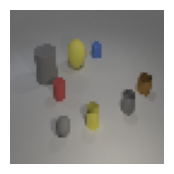}};
    \node[trap,below=2mm of ip3,rotate=180,anchor=center] (e3) {\rotatebox{180}{E}};
    \node[inner sep=0, below=5mm of e3]
      {\includegraphics[width=0.22\linewidth]{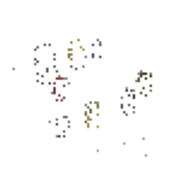}};    
  \end{tikzpicture}
  \vspace*{-\baselineskip}
  \caption{(top) The \ppsvae\ framework. (bottom) Examples of meaningful context points induced by the encoder.}
  \label{fig:hook}
\end{figure}

Conditional neural processes \citep[CNPs]{pmlr-v80-garnelo18a} are a family of models that learn distribution over functions.
In contrast to conventional approaches such as Gaussian processes, which are effective but become computationally expensive once the data size increases, CNPs are both flexible regarding the functions they approximate, thanks to being neural networks, and scalable to large datasets.
In the visual domain, they have been used for contextual image completion.
Given a context set, a set of ordered pairs---observed pixel values and their image coordinates---CNPs learn to impute the other, unobserved, pixels.

While prior work on CNPs primarily focusses on model choices such as inductive biases that allow capturing various properties of the context set better \citep{Gordon2020Convolutional} or dependencies between the unobserved pixel values \citep{DBLP:journals/corr/abs-1807-01622}, we explore a key \emph{dual} question---regarding the context set itself.
Where the context set is typically chosen at random, or derived from some simple statistic (e.g.
pixel variance) to train the CNP, we ask: \emph{which pixels would a CNP like to observe?}
Do such pixels allow better-fitting of CNPs, and do they tell us something meaningful about the underlying image?
We explore these questions from the frame of representation learning, where the context can be viewed as \emph{latent} representations of the image---one that happens to exist in the data space.

From a purely representation-learning perspective, one can relate the question above with that of learning (a) a discrete feature selector as in Concrete Autoencoder \citep[CAE]{conc-autoencoder} and (b) a discrete latent `code', as first established in \citep[VQ-VAE]{10.5555/3295222.3295378}, and subsequently popularised by approaches like DALL-E~\citep{pmlr-v139-ramesh21a}.
Where the CAE employs a global feature selector, we approximate a posterior distribution and where the VQ-VAE learns an arbitrary code, we learn one that directly corresponds the pixels
in the image and is sufficiently expressive to capture image content---measured through reconstruction.
Given the interpretation of our model as \emph{imputing} the remainder of the observation from the given pixel `codebook', we bring together the ideas of discrete representation learning and learning-to-learn stochastic processes (CNPs) into a single framework---the partial pixel specification variational autoencoder (\ppsvae, shown in~\cref{fig:hook}).

Specifically in this work, we
\begin{compactitem}
\item develop an amortised variational inference framework (\ppsvae) to \emph{learn to predict} context points that a CNP can faithfully complete (\cref{sec:model}),
\item provide evidence that learning context along with the CNP learns a better model over images (\cref{sec: log_marginal}),
\item demonstrate that the \ppsvae\ encodes useful and meaningful information in the \emph{learnt} context set---evaluated through both qualitative observation and a classification-probe task --- both in-distribution and out-of-distribution settings (\cref{section:qunat_prob}), and
\item highlight the utility, flexibility, and scalability of \ppsvae\ with improved performance using simple post-hoc augmentations such as dynamic resizing of context sets and reconfiguration of context sets as tiles (\cref{section:qunat_prob}).

\end{compactitem}

\section{Model}
\label{sec:model}

\paragraph{CNPs.}
Given function \(f: \X \to \Y\) mapping observations~\(x \in \X\) to targets~\(y \in \Y\), and \emph{context set} \(\C = \{(x_m, y_m)\}_{m=1}^M\), a CNP \citep{pmlr-v80-garnelo18a} learns a distribution over functions~\(f(x; \C)\)---predicting targets conditioned on context~\(\C\).
For unseen~\(\xt = \{x_t\}_{t=1}^T\), the CNP defines the distribution over~\(\yt = \{y_t\}_{t=1}^T\) as
\begin{tcolorbox}[ams align, colback=white, title=Eq. 1 - CNP's Predictive Distribution,
  top=2pt, bottom=2pt]
{\stepcounter{equation}}
  p_\theta(\yt \mid \xt, C)
  &= \prod_{t=1}^T  \N(y_t \mid \mu_t, \sigma_t) \nonumber\\
  \mu_t, \sigma_t
  &= s_{\theta}(x_t, r_{\theta}(\C)). \nonumber
\end{tcolorbox}

Crucially, it relies on transforming the entire context set~\(\C\) in a permutation-invariant fashion \citep[DeepSet]{NIPS2017_f22e4747} using \(r_{\theta}\), to construct the parameters of the distribution through \(s_{\theta}\), using neural networks as parameters.

\begin{figure}[t!]
  \centering
  \begin{tikzpicture}[thick, bnd/.style={draw,thick,dashed,gray!50,rounded corners=1ex,inner sep=1.5ex}]
    \matrix[column sep=1cm, row sep=8mm] {%
      \node[lat] (a) {\(\bm{a}\)};       & \\
      \node[lat] (xm) {\(\xm\)};    & \node[lat] (ym) {\(\ym\)}; \\
      \node[det] (xt) {\(\xt\)};    & \node[lat] (yt) {\(\yt\)}; \\
    };
    \graph {%
      (xm) -> {(ym), (xt), (yt)};
      {(xt), (ym)} -> (yt);
      (a) -> {(xm), (ym)}
    };
    
    \begin{scope}[on background layer]
      \node[fit=(xm) (xt) (ym) (yt), fill=yellow]{};
    \end{scope}
    \begin{scope}[on background layer]
      \node[bnd, color=black, fit=(yt) (ym),label={[label distance=-1.2em]-75:\(\bm{y}\)}] {};
    \end{scope}
  \end{tikzpicture}
  \;
  \begin{tikzpicture}[thick]
    \matrix[column sep=1cm, row sep=8mm] {%
      \node[ob] (y) {\(\bm{y}\)};        & \\
      \node[lat] (xm) {\(\xm\)}; & \node[det,fill=gray!30] (ym) {\(\ym\)}; \\
       \node[lat] (a) {\(\bm{a}\)};        & \\
    };
    \graph {%
      (y) -> { (xm), (ym) };
      (xm) -> (ym);
      {(xm), (ym)} -> (a);
    };
  \end{tikzpicture}
  \caption{CNP generative model (left yellow); \ppsvae\ generative (left) and inference (right) models.}
  \label{fig:ppsvae-gm}
\end{figure}
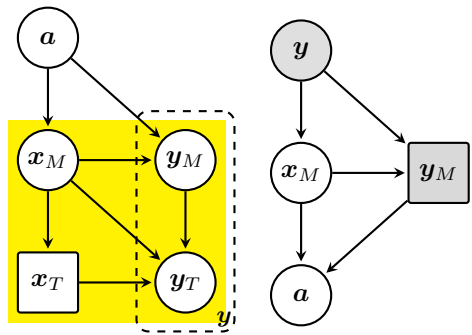

In the image domain, a CNP learns to predict the colour values~\(\yt\) at unseen locations~\(\xt\) given a set of observed pixel locations~\(\xm\) and their corresponding values~\(\ym\).
By observing some small, sparse subset of the image itself, the task here is to impute the rest of the image.
%
Note that, in this setting, knowing the set of observed locations~\(\xm\) implies knowing the set of unseen locations~\(\xt\), as for images of fixed size, one is the complement (\(\xt = {\xm}'\)) of the other.
Learning a CNP in this setting involves (random) sampling of different context sets and subsequent imputation of the values at unseen locations, across a dataset of images.

\paragraph{\ppsvae.}
To answer our question of what kinds of context the CNP would like to observe, and how meaningful this context is, we first cast the CNP as a fully generative model as shown in \cref{fig:ppsvae-gm} (left---yellow area),
\begin{tcolorbox}[ams align, colback=white, title= Eq. 2 - CNP's Generative Model, label=equation: cnp_generative_model, top=2pt, bottom=2pt]
{\stepcounter{equation}}
  \hspace*{-2.5ex}p_\theta(\bm{x}, \bm{y}| M)
  &\!=\! p_\theta(\xm\!)\; p_\theta(\ym\! | \xm\!)\; p_\theta(\yt\! | \xt, \xm, \ym\!) \nonumber
\end{tcolorbox}
%
Here, \(M\) is taken to be a given fixed value, \(p_\theta(\xm)\) defines a distribution over \emph{arrangements} of~\(M\) pixel locations in an image, and \(p_\theta(\ym | \xm)\) a distribution over values at the given locations.
The model can be viewed as generating data in two stages (autoregressive): first generating the values corresponding to the context points, and subsequently, conditioning on these locations and values to impute the values elsewhere on the image. From this, to get to the full \ppsvae\ generative model, we additionally introduce an \emph{abstractive} latent variable~\(\bm{a}\)\footnote{The parameter $\theta$ of the \(p_\theta(\ym|\xm)\) distribution is shared among all data instances. Given that the distribution of values in a pixel $\ym$ can vary enormously depending  on the (both global and local) arrangements of $\xm$, the model will typically struggle to faithfully learn such a distribution across all data instances. We tackle this issue by introducing an \emph{abstractive} latent variable~\(\bm{a}\).} as shown in Figure \ref{fig:ppsvae-gm} (left). The latent variable~\(\bm{a}\) acts as an abstraction of the context set/PPS, providing smooth control over different arrangements and values, while also allowing the model to flexibly learn the mapping between arrangement of pixel locations and corresponding pixel vales.
The full \ppsvae\ generative model can thus be defined as
\begin{tcolorbox}[ams align*, colback=white, title=Eq. 3 - PPS-VAE: Generative Model, top=2pt, bottom=2pt]
{\stepcounter{equation}}
&\hspace*{-1.7ex}p_\theta(\bm{a}, \bm{x}, \bm{y}| M)\\
&\hspace*{-2.7ex}\!=\! p_\theta(\bm{a})\; p_\theta(\xm | \bm{a})\; p_\theta(\ym | \xm, \bm{a})\; p_\theta(\yt | \xt, \xm, \ym)
\end{tcolorbox}\label{eq:ppsvae-p} 
\begin{align*}
  \underset{\text{\colorbox{cyan}{abstract.}}}{p_\theta(\bm{a})}
  &= \N(\bm{a} | \bm{0}, \bm{1}) \\
\underset{\text{\colorbox{cyan}{locations}}}{p_\theta(\xm | \bm{a})}
  &= \prod_{m=1}^M \mathop{GS}(x_m | g_\theta^1(\bm{a})) \\
  \underset{\text{\colorbox{cyan}{pixel values}}}{p_\theta(\ym | \xm, \bm{a})}
  &= \prod_{m=1}^M \N(y_m | g_\theta^2(\xm, \bm{a})) \\
  \underset{\text{\colorbox{cyan}{pixel values}}}{p_\theta(\yt | \xt, \xm, \ym)}
  &= \prod_{t=1}^T \N(y_t | g_\theta^3(\xt, \xm, \ym))
\end{align*}
where \(g_\theta^1, g_\theta^2, \text{and} g_\theta^3\) are parametrised neural networks that transform input values to corresponding distribution parameters, and \(\mathop{GS}\) is the Gumbel-Softmax distribution \citep{maddison2017the, jang2017categorical} which provides a continuous relaxation of the discrete distribution---enabling reparametrised gradient estimation.

The standard CNP formulation estimates the marginal \(p_\theta(\bm{y}|M)\) by sampling uniformly at random from \(p(\xm)\).
One can instead construct a more informative importance-sampled estimator by employing a variational posterior \(q_\phi(\xm |\bm{y}, M)\) in the vein of \citet[VAE]{kingma2014autoencoding}.

Crucially, given a means to generate locations~\(\xm\), one can simply lookup the image~\(\bm{y}\) at those locations to derive~\(\ym\)---an observation itself---as shown in \cref{fig:ppsvae-gm} (right).
From a representation-learning perspective, the context set can be seen as a \emph{partial pixel specification} (PPS) of the image.
The corresponding inference model is
\begin{tcolorbox}[ams align*, colback=white, title=Eq. 4 - PPS-VAE: Inference Model, top=2pt, bottom=2pt]
{\stepcounter{equation}}
  q_\phi(\bm{a}, \xm |\bm{y}, M) = q_\phi(\xm | \bm{y})\; q_\phi(\bm{a} | \xm, \ym)
\end{tcolorbox}
\setcounter{equation}{\theequation-1}
\begin{subequations}
\vspace*{-1.5\baselineskip}
\begin{align}
  \underset{\text{\colorbox{cyan}{locations}}}{q_\phi(\xm | \bm{y})}
  &= \prod_{m=1}^M \mathop{GS}(x_m | h_\phi^1(y, x_{<m})) \label{eq:ppsvae-qxm}\\
  \underset{\text{\colorbox{cyan}{abstract.}}}{q_\phi(\bm{a} | \xm, \ym)}
  &= \N(\bm{a} | h_\phi^2(\xm, \ym)),
    \nonumber
\end{align}    
\end{subequations}
where the generative model independently factorisation \(p_\theta(\xm|\bm{a})\), and the posterior uses an autoregressive formulation.
Again, \(h_\phi^1  \text{and}\ h_\phi^2\) are parametrised neural networks that transform inputs to distribution parameters.
In eq. \ref{eq:ppsvae-qxm}, \({x_{<m}}\) for \(m=1\) is assumed to be null.

Putting the generative and inference models together, we construct the variational evidence lower bound (ELBO) as
\begin{equation}
\log\ p_\theta(\bm{y}|M) \nonumber \geq \E_{q_\phi(\bm{a}, \xm |\bm{y}, M)} \left[
     \log \frac{p_\theta(\bm{a}, \bm{x}, \bm{y}| M)}{q_\phi(\bm{a}, \xm |\bm{y}, M)} \right],
\label{eq:ppsvae-elbo} 
\end{equation}
which can be further expanded as

\begin{tcolorbox}[ams align, colback=white, title=Eq. 5 - PPS-VAE: ELBO, top=2pt, bottom=2pt]
    & \hspace*{-1.7ex}\E_{q_\phi(\bm{a}, \xm |\bm{y})}\nonumber\hspace*{0.1ex}\bigl[\log p_\theta(\yt | \xt, \xm, \ym)p_\theta(\ym | \xm, \bm{a}) \bigr] -\\
    & \hspace*{16ex} -\KL{q_\phi(\bm{a}, \xm | \bm{y})}{p_\theta(\bm{a}, \xm)} \nonumber
\label{eq:ppsvae-elbo-expand} 
\end{tcolorbox}

where~\(\ym\) and~\(\yt\) are observations derived as \(\bm{y} \odot \xm\) and \(\bm{y} \odot \xt\) respectively---lookups for complementary sets of pixel locations.
Note that the abstractive latent~\(\bm{a}\) is reversed in the generative vs. inference models---a location~\(x_m\) sampled from the posterior can only be scored in the generative model once the corresponding~\(\bm{a}\) has been sampled.
This ensures that the complex transformation involved in~\(\xm \to \ym\) is captured by the abstractive latent.

\paragraph{Inductive biases.}
In the first instance, given our focus on the visual domain, we employ a specific variant of CNPs called the ConvCNP \citep{Gordon2020Convolutional}, which explicitly incorporates translation equivariance and locality constraints enforced by convolutional neural network (CNN) filters.
We use this same inductive bias with CNNs in the inference model \(q_\phi(\xm |\bm{y})\).
We find this to be an important design decision, as attempting to model these components using the standard multi-layer perceptron (MLP) \citet[as in][]{pmlr-v80-garnelo18a} causes issues, primarily with the model using the context set/PPS as a generic lookup table, with little to no spatial meaning (see Appendix \ref{append:mlp_ind_bias}).
The CNN-based setup provides the requisite inductive bias that allows meaningful spatial arrangement of points (see \cref{section: vis-inspection}).

\begin{figure*}[t!]
\subcaptionbox{\label{figure: pps_clevr_m_30}CLEVR, $M=32$}{\includegraphics[width=0.5\textwidth]{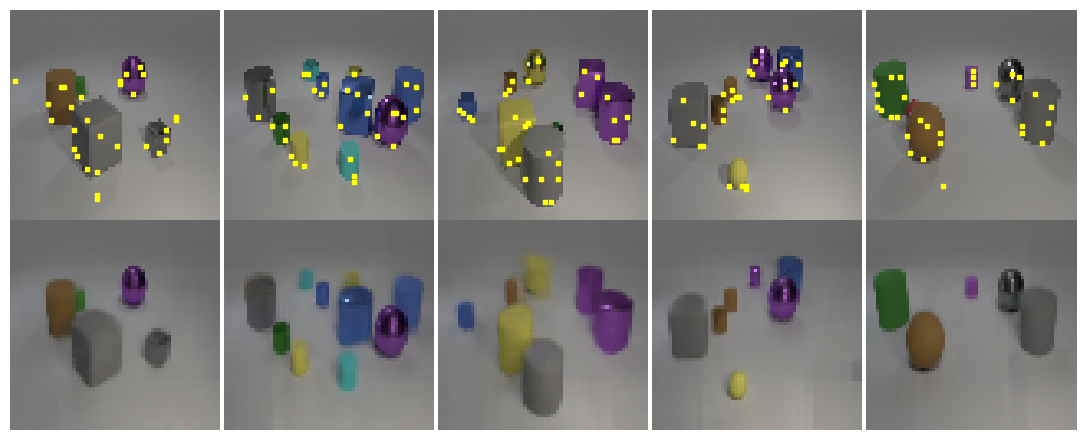}}
\subcaptionbox{\label{figure: pps_clevr_m_128}CLEVR, $M=128$}
{\includegraphics[width=0.5\textwidth]{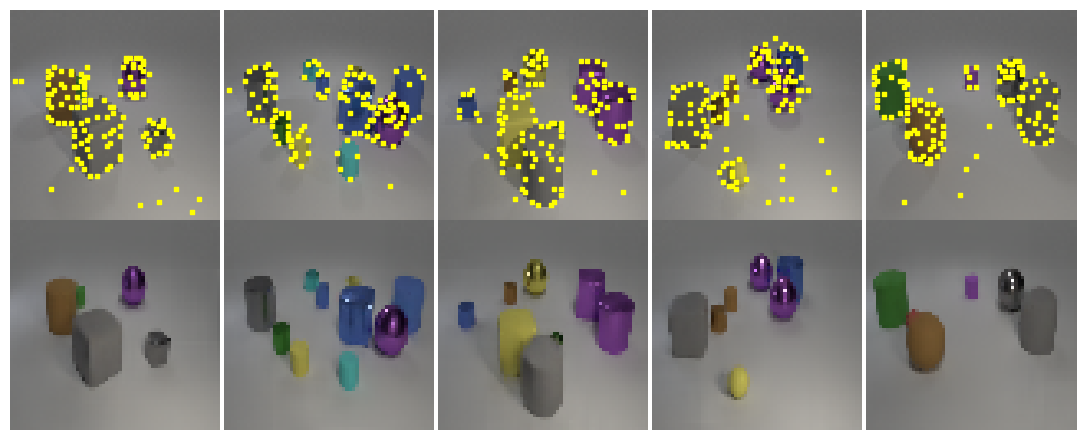}}
\subcaptionbox{\label{figure: pps_c10_m_15}CelA, $M=32$}
{\includegraphics[width=0.5\textwidth]{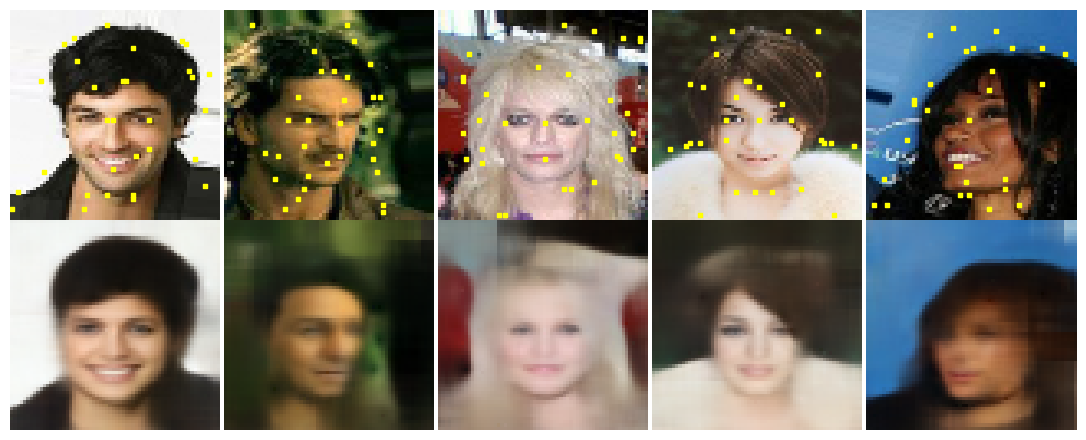}}
\subcaptionbox{\label{figure: pps_c10_m_60}CelA, $M=128$}
{\includegraphics[width=0.5\textwidth]{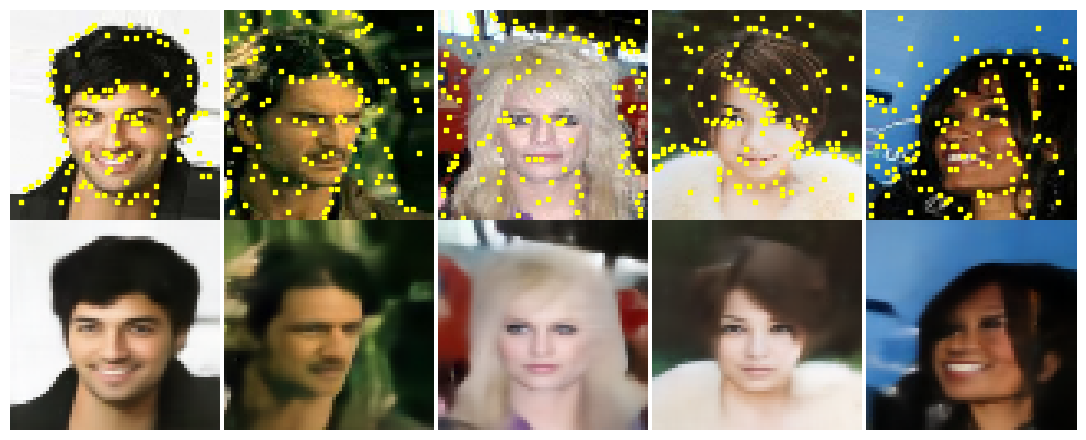}}
\subcaptionbox{\label{figure: pps_ca_m_15}t-ImageNet, $M=32$}
{\includegraphics[width=0.5\textwidth]{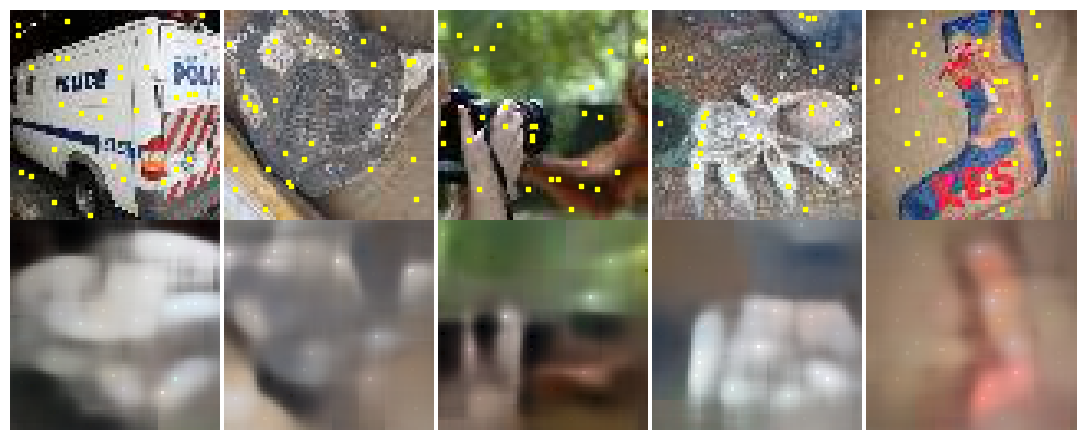}}
\subcaptionbox{\label{figure: pps_ca_m_30}t-ImageNet, $M=128$}{\includegraphics[width=0.5\textwidth]{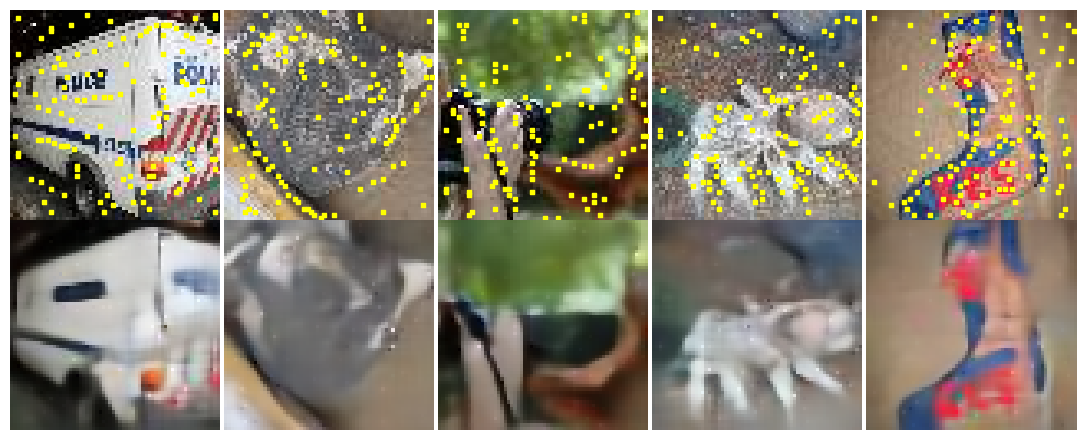}}
\vspace*{-1.5ex}
\caption{Visualisation of the spatial arrangement of the context set for \ppsvae\ on three datasets (test images): CLEVR (a,b) and CelA (c,d) and t-ImageNet (f,e).  In each figure [a-f] the first row corresponds to the original image, together with the inferred context set denoted by the yellow squares. The second row corresponds to the reconstructed images.}
\label{figure: pps_vis}
\end{figure*}

\vspace*{-0.5\baselineskip}
\section{Experiments}
\vspace*{-0.5\baselineskip}
Our primary goal here is to understand properties of the context set/PPS.
For this we:
\begin{compactitem}
\item Estimate the log marginal distribution to understand if the learned (rather than randomly
  sampled) context set helps better model the images (Subsection \ref{sec: log_marginal}),
\item Analyse the kinds of points the model chooses; 1-to-1 correspondence between the PPS and an image allows us to perform a visual inspection (see Subsection \ref{section: vis-inspection}),
\item  Quantify how representative the context set is of the object classes.  We do this through the lens of classification, by probing the context set on: 1) in-distribution ---\ppsvae\ pre-training dataset and the classification dataset are the same, 2) out-of-distribution (ood) datasets---pre-training dataset differs from the classification dataset (see Subsection \ref{section:qunat_prob}). Moreover we discuss \textbf{an ability of the PPS encoder to change capacity during inference}, and
\item Demonstrate flexibility and scalability through larger images and ood reconstruction (see Subsection \ref{section: misc-prop}).
\end{compactitem}

\paragraph{Datasets.} We use four standard vision datasets: FER2013 \citep{fer2013}, CelebA \citep[CelA]{liu2015faceattributes}, CLEVR \citep{johnson2017clevr} and Tiny Imagenet \citep[t-ImageNet]{tiny-imagenet} with resolution at 64x64.

\begin{table}[b!]
  \setlength{\tabcolsep}{3pt}
  \centering
  \caption{Estimated $\log p_\theta(\bm{y}|M) (\uparrow)$ with 800 samples. For all models $M=128$. \fcolorbox{black}[HTML]{a8ddb5}{-} the best performance. }
  \label{table:elbo}
  \begin{tabular}{@{}lcccc@{}}
    \toprule
    &  FER2013 &  CelA &  CLEVR &t-ImageNet\\
    \midrule
    \ppsvae  &  \cellcolor[HTML]{a8ddb5} 4951 &  \cellcolor[HTML]{a8ddb5}14210 &  \cellcolor[HTML]{a8ddb5}16611&  \cellcolor[HTML]{a8ddb5}16324 \\
    PPS-CAE &  4471 &  12162 &  16089 &  15832 \\
    ConvCNP  &  4472 &  12064 &  15981 &  15793\\
    \bottomrule
  \end{tabular}
\end{table}

\paragraph{Models.}
For all datasets we train \ppsvae\ with \(M = \{32, 64, 128 \} \). Where concerned with performance on a  metric, \ppsvae\ with $M = 128$ perform the best, results for other $M$'s are in Appendix \ref{append:class_perfm_vs_M}.
To better ground the experimental results, we employ three baselines: VQ-VAE~\citep{10.5555/3295222.3295378}, FSQ-VAE~\citep{mentzer2023finite} and PPS-CAE a variant of CAE~\citep{conc-autoencoder} where we use the same encoder and set ConvCNP~\citep{Gordon2020Convolutional} as a decoder.
Also, in \cref{sec: log_marginal} we fit ConvCNP with the random selection of points and in Subsection \ref{section:qunat_prob} we use RAND-PPS model --- an encoder that randomly samples $M$ points from an image.
We train the models once and use them in all the experiments (details in Appendix \ref{append:imp_and_train_details}).

\subsection{Model Fit}
\label{sec: log_marginal}
\vspace*{-0.5ex}

Here we estimate $\log p_\theta(\bm{y} | M)$ (see Appendix \ref{append:log_marg_est}) and use it to compare the models (see Table \ref{table:elbo}).
The first observation is that PPS-CAE outperforms ConvCNP on all dataset but one, which provides evidence that learning context set helps modelling distribution over the
images. The second observation is that learning posterior of the context set (\ppsvae) instead of just a prior (PPS-CAE) provides a further improvement. In Appendix \ref{append:log_marginal_vs_M}, for \ppsvae, we further  estimate $\log p_\theta(\bm{y} | M)$ for various values of $M$ and find that for all datasets, increasing $M$ results in better performance. 

{\color{darkgray}\emph{\textbf{Findings:} Learning (instead of randomly
sampling) context set  helps modelling distribution over the images.}}

  %

\begin{table*}[t!]
  \setlength{\tabcolsep}{2.5pt}
  \centering
  \caption{Object classification (in-distribution): Classifiers trained over three seeds with early stopping, reporting mean F1-macro scores. A:13---Chubby, A:20---Male, A:25---Oval Face.  \fcolorbox{black}[HTML]{e0f3db}{- $\pm$ -} best performance among models with an encoder; \fcolorbox{black}[HTML]{a8ddb5}{- $\pm$ -} absolute best performance. 128$\rightarrow$256 means the model was trained on $M=128$ and evaluated with $M=256$.}
  \label{table:in-domain-class}
  \scalebox{1.}{%
  \begin{tabular}{@{}l@{\quad}lcccccc@{}}
    \toprule
    & &   CelA (A:13) &  CelA (A:20) &  CelA (A:25)  & FER2013 &  CLEVR & t-ImageNet  \\
    \midrule
    \multirow{6}{*}{\rotatebox[origin=c]{90}{\;\small \textsc{Baselines}}}
    & PPS-RAND (points) & 60.92 $\pm$ 1.28 &  90.89 $\pm$ 0.06  & 56.20 $\pm$ 0.87 &   34.97 $\pm$ 0.38 &  36.17 $\pm$ 3.39  &  21.86 $\pm$ 0.31    \\
    & PPS-RAND (post-hoc tiles) & 66.91 $\pm$ 1.01 &  95.10 $\pm$ 0.16  & 60.09 $\pm$ 0.24 &   43.30 $\pm$ 0.43 &  63.20 $\pm$ 0.89  &  33.52 $\pm$ 0.23    \\
    & PPS-CAE (points) & 61.29 $\pm$ 0.91  & 91.35 $\pm$ 0.14 &  58.36 $\pm$ 0.45  &  35.28 $\pm$ 0.50  &  48.50 $\pm$ 2.77  &  22.53 $\pm$ 0.30   \\
    & PPS-CAE (post-hoc tiles) & 67.16 $\pm$ 0.90  & 95.46 $\pm$ 0.07 &  60.75 $\pm$ 0.67  &  44.32 $\pm$ 0.66  &  74.85 $\pm$ 0.49  &  33.55 $\pm$ 0.20  \\
    & VQ-VAE & 68.59 $\pm$ 0.04  & 94.83 $\pm$ 0.13 &  62.44 $\pm$ 0.34  &  50.98 $\pm$ 0.52  &  75.91 $\pm$ 0.47  &  29.02 $\pm$ 0.08   \\
    
     & FSQ-VAE & 68.19 $\pm$ 0.81  & 95.21 $\pm$ 0.11 &  62.28 $\pm$ 0.22   &  45.46 $\pm$ 0.15 &  73.27 $\pm$ 0.36   &  31.03 $\pm$ 0.40  \\
     \midrule
     \multirow{3}{*}{\rotatebox[origin=c]{90}{\;\small \textsc{Our}}}
    & \ppsvae\ (points) & 69.00 $\pm$ 0.38 &  94.86 $\pm$ 0.12 & 62.13 $\pm$ 0.50  &  46.72 $\pm$ 0.62  &  90.21 $\pm$ 0.28 &  29.56 $\pm$ 0.27    \\
    & \ppsvae\ (points) 128$\rightarrow$256 & 69.94 $\pm$ 0.50 &  95.70 $\pm$ 0.07 & 62.02 $\pm$ 0.50  &  \cellcolor[HTML]{e0f3db}51.61 $\pm$ 0.57  &  93.38 $\pm$ 0.64 &  33.93 $\pm$ 0.16    \\
    & \ppsvae\ (post-hoc tiles) & \cellcolor[HTML]{e0f3db}70.94 $\pm$ 0.09 & \cellcolor[HTML]{e0f3db} 96.21 $\pm$ 0.04 & \cellcolor[HTML]{e0f3db}62.94 $\pm$ 0.10  &  49.38 $\pm$ 0.39  & \cellcolor[HTML]{a8ddb5} 94.62 $\pm$ 0.28 & \cellcolor[HTML]{e0f3db} 35.00 $\pm$ 0.04    \\
    \midrule
     & Image   & \cellcolor[HTML]{a8ddb5}73.47 $\pm$ 0.49  &   \cellcolor[HTML]{a8ddb5}97.55 $\pm$ 0.02 &\cellcolor[HTML]{a8ddb5} 64.49 $\pm$ 0.25  &  \cellcolor[HTML]{a8ddb5}61.56 $\pm$ 0.17  &  91.90 $\pm$ 0.30  &  \cellcolor[HTML]{a8ddb5}43.68 $\pm$ 0.03     \\
    \bottomrule
  \end{tabular}}
\end{table*}

\vspace*{-1ex}
\subsection{Visual Inspection of PPS}
\label{section: vis-inspection}
\vspace*{-0.5ex}

Since there is 1-to-1 correspondence between pixels in the context set an the original image it allows us to perform a qualitative observation of the chosen pixels and put forward hypothesis regarding how PPS-VAE abstracts information for different settings of~\(M\). Results are shown in \cref{figure: pps_vis}, with additional examples given in Appendix \ref{append:vis_rec_imgs}.\@ 

The patterns that context sets form can be summarised with the following observations: (1) boundary points between objects and the background generally describe shape, (2) points on the object can capture `interior’ colour, and part locations and (3) background points capture complexity outside the objects (e.g. uniform colour etc.).

We also emphasise that these patterns are more pronounced when $M$ is sufficiently large (e.g. $M=128$). However, when $M$ relatively small compared to the complexity of an image, the context set appears scattered---possibly because the model tries to ``cover'' the complexity of the image, by exploring the image space rather than exploiting any region; the former is likely to reconstruct the \emph{whole} image better.

{\color{darkgray}\emph{\textbf{Findings:} The analysis shows that, when $M$ is sufficiently large,  the context set forms  pronounced patterns with the following three types of points: boundary points around objects, points inside an object and background points.}}

\begin{table*}[t!]
  \setlength{\tabcolsep}{5pt}
  \centering
  \caption{Object classification --- out-distribution setting. Classifiers trained over three seeds with early stopping, reporting mean F1-macro scores. A:13 --- Chubby, A:20 --- Male, A:25 --- Oval Face. {\color{orange}- $\pm$ -} in distribution encoders (copied from \cref{table:in-domain-class}); \fcolorbox{black}[HTML]{a8ddb5}{- $\pm$ -} the best performance. All PPS based models are evaluated on points. } 
  \label{table:out-dist-class}
  \begin{tabular}{@{}l@{\quad}lccccc@{}}
    \toprule
    &  &  CelA (A:13)  &  CelA (A:20) &  CelA (A:25) &  CLEVR &  t-ImageNet   \\
    \midrule
     \multirow{3}{*}{\rotatebox[origin=c]{90}{\scriptsize \textsc{PPS-VAE}}}
    & CelA &   {\color{orange}69.00 $\pm$ 0.38} &  {\color{orange}94.86 $\pm$ 0.12} & {\color{orange}62.13 $\pm$ 0.50} & 80.27 $\pm$ 1.06  & \cellcolor[HTML]{a8ddb5}29.59 $\pm$ 0.25    \\
    & CLEVR   &   67.02 $\pm$ 0.38  &  93.39 $\pm$ 0.09 &  60.33 $\pm$ 0.25  &   {\color{orange}90.21 $\pm$ 0.28}  &  25.05 $\pm$ 0.22   \\
    & t-ImageNet  &  67.09 $\pm$ 0.34   &   93.68 $\pm$ 0.14 &  61.28 $\pm$ 0.40 &  \cellcolor[HTML]{a8ddb5}80.66 $\pm$ 0.59  & {\color{orange}29.56 $\pm$ 0.27}    \\
    \midrule
     \multirow{3}{*}{\rotatebox[origin=c]{90}{\scriptsize \textsc{FSQ-VAE}}}
    & CelA &   {\color{orange}68.19 $\pm$ 0.81}  & {\color{orange}95.21 $\pm$ 0.11} &  {\color{orange}62.28 $\pm$ 0.22} & 69.07 $\pm$ 0.63  &  29.24 $\pm$ 0.07    \\
    & CLEVR   &   69.21 $\pm$ 0.88  &  94.90 $\pm$ 0.03 &  62.58 $\pm$ 0.25  & {\color{orange}73.27 $\pm$ 0.36}  &  28.48 $\pm$ 0.37   \\
    & t-ImageNet  &  \cellcolor[HTML]{a8ddb5}70.04 $\pm$ 0.24    &   \cellcolor[HTML]{a8ddb5}95.07 $\pm$ 0.02 &  \cellcolor[HTML]{a8ddb5}62.87 $\pm$ 0.24 &  69.35 $\pm$ 0.66  &   {\color{orange}31.03 $\pm$ 0.40}     \\
    \midrule
     \multirow{3}{*}{\rotatebox[origin=c]{90}{\scriptsize \textsc{VQ-VAE}}}
    & CelA &   {\color{orange}68.59 $\pm$ 0.04}  & {\color{orange}94.83 $\pm$ 0.13} &  {\color{orange}62.44 $\pm$ 0.34}   &  68.22 $\pm$ 0.31 & 28.56 $\pm$ 0.26   \\
    & CLEVR   &   66.28 $\pm$ 0.57  &  92.93 $\pm$ 0.12 &  60.82 $\pm$ 0.27  &  {\color{orange}75.91 $\pm$ 0.47}  &  24.16 $\pm$ 0.22   \\
    & t-ImageNet  &  68.92 $\pm$ 0.24    &   94.40 $\pm$ 0.07 &  62.44 $\pm$ 0.13 &  68.26 $\pm$ 0.32  &  {\color{orange}29.02 $\pm$ 0.08}     \\
    \midrule
     \multirow{3}{*}{\rotatebox[origin=c]{90}{\scriptsize \textsc{PPS-CAE}}}
    & CelA &   {\color{orange}61.29 $\pm$ 0.91}  & {\color{orange}91.35 $\pm$ 0.14} &  {\color{orange}58.36 $\pm$ 0.45}  & 37.22 $\pm$ 2.53  &  23.00 $\pm$ 0.64   \\
    & CLEVR   &   61.59 $\pm$ 0.50  &  91.77 $\pm$ 0.05 &  58.09 $\pm$ 0.50  &  {\color{orange}48.50 $\pm$ 2.77}  &  23.82 $\pm$ 0.08   \\
    & t-ImageNet  &  61.80 $\pm$ 0.91    &   90.78 $\pm$ 0.08 &  58.29 $\pm$ 0.46 &  37.22 $\pm$ 0.34  &  {\color{orange}22.53 $\pm$ 0.30}     \\
    \bottomrule
  \end{tabular}
\vspace*{2ex}
\end{table*}

\subsection{Quantitative Analysis: PPS Probing}
\label{section:qunat_prob}

Having observed  that the context sets/PPS do indeed appear to capture meaningful features, we conduct further analyses to quantify how meaningful they can be.
We do this through the lens of classification, by probing the context set/PPS (\(\ym\)) (in in-distribution and out-of-distribution settings) to see how well it captures class-relevant information.
Note that we simply use this as a mechanism to evaluate how well the model captures class-specific information; we do not attempt to engineer a SOTA classifier.

\paragraph{PPS.}
To evaluate the utility of the context set/PPS \(\ym\) suffices (also referred to as points). 
Using the location variable~\(\xm\) did not provide further benefit. For all the datasets we set $M=128$, which is $\approx$ 3.13$\%$ of the original number of pixels.
As an additional experiment, we augment $\ym$ at inference time by adding to each pixel in $\ym$ 8 neighbouring pixels---creating 3x3 tiles after pre-training. We call these post-hoc tiles. This achieves two things: (1) increase the amount of information in the latent without re-training the model and (2) test if the points in $\ym$ represent content well enough for a task and if surrounding points help. This augmentation increases the size of PPS to  $\approx$ 28.13$\%$ of the original number of pixels.

\paragraph{Baselines.}
The first baseline employs the whole image~\(\bm{y}\) (denoted Image), and is used as a yardstick to see how well a restricted context set does.
The second baseline employs a random selection of context points~\(\ym\) (denoted PPS-RAND) to provide contrast against a more informative selection of context set.
Given the discussion in \cref{sec:related_work} of how the FSQ/VQ-VAE can be seen as a selective codebook, but without spatial meaning, we employ it as an additional baseline to see how the constraint of spatial relevance affects classification. Finally, we use PPS-CAE to benchmark global vs instance-specific context sets. 

\paragraph{Classification Tasks.} The datasets we chose for the pre-training of PPS-VAE and the baseline models come with associated classification tasks. Such that t-ImageNet comes with labels of 200 different classes, FER2013 associate each facial expression with one of the seventh emotion categories, CLEVR comes with labels for number of objects in an image and finally CelebA includes 40 binary attributes associated with facial characteristic. We select 3 generic attributes: A:13 --- Chubby, A:20 --- Male, A:25 --- Oval Face. 
 
\paragraph{Classifier.}
As the base classifier, we employ the ConvMixer  \citep{trockman2023patches} architecture, training each instance entirely from scratch.
The encoders of PPS-VAE and the baseline models: VQ-VAE, FSQ-VAE and PPS-CAE are held frozen during the training of the classifier --- only the parameters of ConvMixer are trained. 
%
We \textbf{do not} perform additional data preprocessing or augmentation. This  give us better signal of whether the performance gains are from information encoded in the representations.

\paragraph{In-Distribution vs Out-Of-Distribution Settings.} In the in-distribution setting the data used to pre-train the model and the classifier are the same. In the out-of-distribution setting, we take a pre-trained model over a dataset, say t-ImageNet, and evaluate it on say, the CelA and CLEVR datasets. As before, the encoders stay frozen.

\subsubsection{In-Distribution setting: Results}

Based on \cref{table:in-domain-class} we make the following observations.
\paragraph{PPS vs Baselines.}
First, the arrangements of points inferred by \ppsvae\ is more indicative of the class than of PPS-RAND. This indicates that the model performs abstraction to preserve the information related to class labels.
Second, on average, \ppsvae\  performs on par with the baseline models with the pre-trained encoder: FSQ-VAE, VQ-VAE  on FER2013, t-ImageNet and CelA datasets, while outperforming the models with a large margin on CLEVR.  The performance on CLEVR is associated with identifying a right number of objects, hence the high classification performance  achieved by \ppsvae\ shows that it has potential to represent abstract object information. Also, not surprisingly, context-set learned by empirical prior of PPS-CAE lags behind of PPS-VAE. This is because PPS-VAE allows us to infer an instance specific context-set while PPS-CAE infers global context set, which may lack instance specific information required for the task. Finally, ConvMixer trained on the original images performs the best on average, which isn't too surprising since $\mathbf{y}$ contains the original information, while the baselines and \ppsvae\ learn abstractions which may result in information loss.

\paragraph{Post-hoc Tiles.} When augmented with the pos-hoc tiles representations inferred by \ppsvae\ dominate the baselines only marginally lagging behind VQ-VAE on FER2013.  Moreover, the \ppsvae\ with post-hoc tiles outperforms the Image baseline on the CLEVR dataset. 

\begin{figure}[b!]
  \centering
  \setlength{\tabcolsep}{4pt}
  \begin{tabular}{ccc}
  $128\rightarrow32$  & $128$ & $128\rightarrow256$\\
    \includegraphics[width=0.11\textwidth]{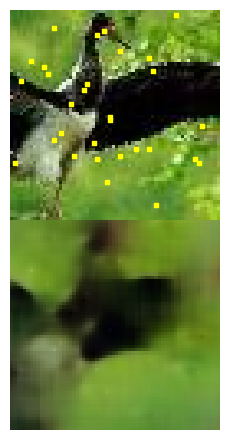} &\includegraphics[width=0.11\textwidth]{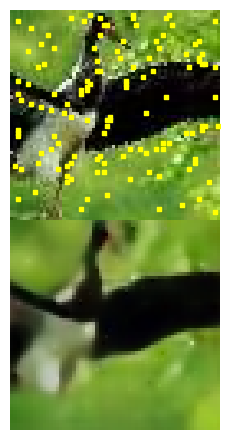}
    & \includegraphics[width=0.11\textwidth]{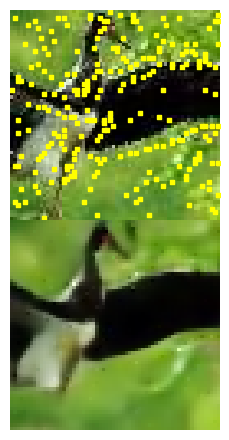} \\
    
  \end{tabular}
  \caption{Visualisation of PPS for changing of $M$ at inference time. \ppsvae\ was pre-trained with $M=128$.}
  \label{figure: pps-m-change-vis}
\end{figure}
\paragraph{Extrapolation of $M$ at Inference.}
A differentiating property of our model is the ability to increase the capacity of the latent representation (PPS) at inference time. We can encode more information in the context set by simply increasing $M$, without retraining the model unlike in the case of VQ-VAE and FSQ-VAE. This can be beneficial in scenarios where a downstream task is complex and $M$ used during the training is not high enough (e.g. due to the computational constraints) to encode all the relevant information to achieve desirable performance on the task. \Cref{figure: pps-m-change-vis} depicts what happens to PPS when the capacity is decreased (left image) or increased (right image).  Performance wise when the capacity is increased $128\rightarrow256$ the classification performance approaches the post-hoc tiles, even allows achieving best performance on FER2013 among the baselines (see Appendix \ref{append:inc_M_capacity}).

\subsubsection{Out-Of-Distribution setting: Results}

We provide results in Table \ref{table:out-dist-class} and make the following observations. First,  PPS-VAE, FSQ-VAE and VQ-VAE still perform strongly compared to the in-distribution setting. Moreover, while PPS-CAE displays slight increase in performance on most of the datasets except CLEVR, for both in-distribution and ood settings classification performance is close to random.  Second for PPS-VAE, FSQ-VAE and VQ-VAE, pre-training on t-ImageNet allows better generalisation to ood images than when pre-trained on the other two datasets. Overall we conclude that the context set learned by the PPS-VAE provides a degree of generalisation, but this varies with the dataset. The same applies to FSQ-VAE and VQ-VAE. We provide additional qualitative observations for PPS-VAE in Appendix \ref{append:vis_out_dist_rec}.

\begin{table}[t!]
  \setlength{\tabcolsep}{3pt}
  \centering
  \caption{ F1-macro scores. Object classification  --- out-of-distribution setting but with trained classifiers (trained in $128\rightarrow256$ Table \ref{table:in-domain-class} experiment).  {\color{orange}- $\pm$ -}  results are  form Table \ref{table:in-domain-class}. A:13---Chubby, A:20---Male, A:25---Oval Face. } 
  \label{table:out-dist-class_m_ext}
  \scalebox{0.98}{%
  \begin{tabular}{@{}l@{\quad}ccc@{}}
    \toprule
    &  CelA  &  CLEVR &  t-ImageNet   \\
    \midrule
    CelA (A:13) &   {\color{orange}69.94 $\pm$ 0.50} &   68.47 $\pm$ 2.39&   67.98 $\pm$ 1.80     \\
    CelA (A:20)    &  {\color{orange}95.70 $\pm$ 0.07}&   95.71 $\pm$ 0.07&   95.71 $\pm$ 0.05   \\
    CelA (A:25)  &  {\color{orange}62.02 $\pm$ 0.50} &   61.11 $\pm$ 0.83&   61.28 $\pm$ 0.96    \\
    CLEVR  &   92.56 $\pm$ 0.40&  {\color{orange}93.38 $\pm$ 0.64} &   92.60 $\pm$ 0.40   \\
    t-ImageNet &   33.80 $\pm$ 0.40 &   33.59 $\pm$ 0.13 &   {\color{orange}33.93 $\pm$ 0.16}    \\
    \bottomrule
  \end{tabular}}
\end{table}

\paragraph{Extrapolation of $M$ at Inference.}
We also test if the increased capacity of PPS can be used in the out-distribution setting. We reuse the pre-trained encoders and the classifiers---trained in the previous Subsection see ( $128\rightarrow256$ Table \ref{table:in-domain-class} experiment).  We report results in \cref{table:out-dist-class_m_ext} and note that these are very close to the in-distribution settings suggesting that increased capacity does not jeopardise out-of-distribution generalisation. The classifiers can be reused with minimal loss in the performance if any.

{\color{darkgray}\emph{\textbf{Findings:} In-distribution: probing reveals that 1) the context set preserves class label information which is on par or better than baselines 2) augmented or increased capacity $\ym$ provides better features for the classifier than the original image $\mathbf{y}$ on CLEVR dataset.
Out-of-distribution: representations learned by \ppsvae, FSQ/VQ-VAE  are robust to out-of-distribution images and can be used with a slight loss of performance on the tasks associated with the images. The degradation of performance depends on the pre-training dataset.}}

\subsection{Miscellaneous Properties} 
\label{section: misc-prop}

\subsubsection{Scalability}
\label{section: scale}
Encoder of PPS-VAE is autoregressive. As with any autoregressive model, a particular bottleneck is its computational complexity, which gets worse with increasing sequence length ($M$). Let $T$ be a computation complexity of a computational block (e.g. CNN) and let the encoder and the decoder is build of the same block. Then the computational complexity will be $\mathcal{O}(M*T)$ assuming $M$ is larger than the number of the blocks in the decoder.  In this section we discuss how to ameliorate this.

\paragraph{Parallel Inference of Points.}
One way to speed up the encoder is to make inference of the points in the context set independent of each other --- inference of all $M$ points in one shot. However, in previous experiments, we found that it results in inferior performance compared to an autoregressive encoder (see Appendix \ref{append:parallel_vs_ind_enc}). Instead, we use mixture of the two --- autoregressive encoder, which at each step infers $K$ points in parallel instead of 1. This reduces complexity to  $\mathcal{O}(M/K*T)$. In our experiments we set $K=8$.

\begin{figure}[t!]
  \centering
  \includegraphics[width=0.15\textwidth]{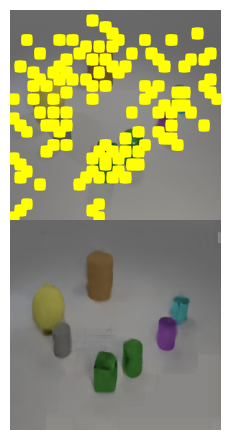}
  \includegraphics[width=0.15\textwidth]{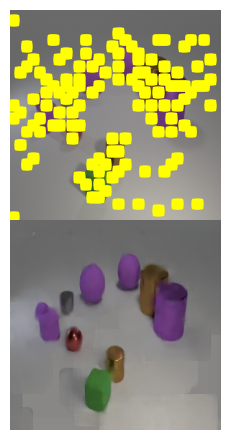}
\includegraphics[width=0.15\textwidth]{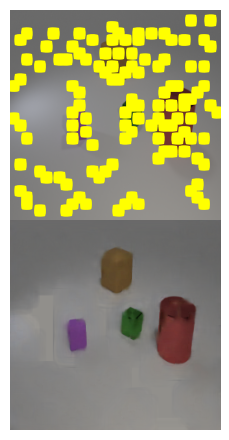}

\caption{Spatial arrangement of the context set for \ppsvae\ tiles. Image size is 256x256, with 8x8 tiles.}
\label{figure: pps_tiles}
\end{figure}

\paragraph{Tiles.}
PPS-VAE is also scalable to large image size. To achieve this we introduce an additional convolutional layer to the encoder that reduces the resolution of an image to specified size - otherwise the model stays the same. For example, given an image of resolution 256x256 the encoder reduces it to 32x32 by producing non-overlapping tiles of size 8x8 (see Figure \ref{figure: pps_tiles}). The decoding is happening in the original resolution 256x256.

\subsubsection{Zero-shot Reconstruction}

Additionally, we test if PPS-VAE can reconstruct an image from an out-of-distribution dataset.  We take a pre-trained model on one of the three datasets and evaluate on the remaining two.  The results can be found in Appendix \ref{append:vis_out_dist_rec}. When  PPS-VAE pre-trained on either CelA or t-ImageNet it can reconstruct images from an out-distribution dataset. For example when trained on CelA it can reconstruct geometric shapes of CLEVR or generic object such as car of t-ImageNet, though with a reduced quality. However, when pre-trained on CLEVR the reconstruction is poor and a lot of artefacts are introduced. The same is observed for FSQ/VQ-VAE (see Appendix \ref{append:vis_out_dist_rec}).

\section{Related Work}
\label{sec:related_work}

CNPs \citep{pmlr-v80-garnelo18a} are a flexible and scalable framework for modelling distributions over functions.
The framework, now more generally referred to as Neural Process Family (NPF) have seen increased popularity, with the different approaches exploring a range of features of the model.
One such approach is the adaption of the CNP to properties of the data
\citep{Gordon2020Convolutional,kawano2021group, DBLP:journals/corr/abs-1812-05212}.
Another approach seeks improved modelling of the output dependencies between function values \citep{DBLP:journals/corr/abs-1807-01622}.
Various other approaches exist; see \citet{arxiv.2209.00517} for an extensive survey.
While all such approaches explore the model's features, to the best of our knowledge, none explore the characteristics of the context set itself.

From a representation-learning perspective, the closest to ours is the VQ-VAE \citep{{10.5555/3295222.3295378}}.
The ability to discretise representation, and learn such a discrete `codebook' through differentiable variational inference that the VQ-VAE employs, has seen successful use in more advance models such as DALL-E~\citep{pmlr-v139-ramesh21a}.
However the types of codebooks that VQ-VAE learns are not interpretable, and it typically needs additional components, such as learning a separate prior, in order to truly function as a generative model over observed data.

The perspective of learning latent representations/features that apply directly on the data domain, can also be compared to work that exposes \emph{attention} mechanisms \citep{Bahdanau2014NeuralMT, 10.5555/2969033.2969073} employed for tasks.
The process of inferring context points can be interpreted as a locally-restrictive way of attending to relevant parts of the image data.
Specifically, such a perspective aligns best with hard attention methods \citep{10.5555/2969033.2969073} as opposed to soft-attention \citep{Bahdanau2014NeuralMT} by virtue of explicitly selecting particular pixels.

Furthermore, inferring context points can also be viewed as a variant of Masked Image Modeling \citep[MIM]{Pathak2016ContextEF}.
MIM involves learning models and representation in a self-supervised fashion by masking parts of an image and attempting to impute them.
More recently, this has been studied extensively as masked autoencoders~\citep[MAE]{he2022masked}.
The imputation task itself is strongly connected to what CNPs do, and one could ask a similar question of MIMs that we ask of CNPs: what kinds of masks do MIMs like to impute?
In fact such a question was indeed asked in work by \citet[ADIOS]{shi2022adversarial} who learnt masks simultaneously with a feature extractor in an adversarial fashion.
This however, is not generative, and as with as masking-based approaches, involves complications with how to specify and generate masks in a sensible manner.
A key distinction is in terms of the sparsity of observed data---MIM and related approaches typically imputes a small part of the image, where CNPs have a more complex task given sparse input.
\ppsvae\ employs context points as weak specifiers of which parts of the image to contextualise, leaving to the CNP itself the question of how to use that specification to capture relevant local and global information from the data.

\section{Discussion}
We present \ppsvae, a novel VAE framework that allows us to infer context set/PPS for conditional neural processes (CNPs).
We formulate our model and evaluate it across multiple vision datasets, while exploring the utility of learning context sets in both unsupervised and supervised manner.
First, we show that the learning distribution over PPS results in better models for images.
Then, we observe that with the appropriate inductive biases and latent variables, the model is able to induce context sets that are visually meaningful. We validate this observation quantitatively through the lens of classification.
On the classification tasks, \ppsvae\ achieves superior performance against the baselines and PPS resulting in better features for a classifier than an original image is on the CLEVR dataset indicating that the framework has promise as a model for learning meaningful representations of data. Additionally, we test our model on the same classification tasks but in out-of-distribution settings showing that it can infer PPS that generalises to an out-of-distribution datasets. Also, we show a differentiating property of PPS-VAE --- an ability to change the capacity of PPS at inference time. Our model, however, has a number of limitation which we would like to outline: 
\begin{compactitem}
    \item Presently we provide an observatory analysis of the induced context set an put forward hypothesis regarding types of points the model learns. However, human level interpretability of the context set is limited. To improve it, instead of inferring a single location, a more interpretable encoder could capture M ‘closed’ regions. This would allow us to compare against the slot-attention models such as \citet{locatello2020objectcentric}.
    \item Exploration of inductive biases, and modelling updates would be interesting avenues to see if the latent variable can capture relevant information more cleanly.
    \item Presently we fix $M$ to a certain value and provide analysis for its various values. However, it may be limiting to decide on the value of $M$ beforehand because we do not know what value would be optimal for each image in a dataset. Allowing the model to decide on the value of $M$ during the learning based on dataset may solve this issue.
\end{compactitem}


\section{Broader Impacts}
The work we describe in this paper aims to improve the interpretability of the latent representations. We foresee that further development of the described algorithm  may allow 
 to overcome a number of drawbacks of Deep Generative Models (DGMs) e.g. \citet{feng2023trainingfree, conwell2022testing} by addressing these issues at the representation level (e.g. by explicit manipulation of latent representations to rectify mistakes DGMs). However, the current work is  algorithmic in nature. And at present stage is not tied to particular applications, let alone deployments.


\section*{Acknowledgements}
The authors gratefully acknowledge the support of ELIAI (The EdinburghLaboratory for Integrated Artificial Intelligence) EPSRC (grant no EP/W002876/1). Ivan Titov also acknowledges Dutch National Science Foundation (NWO Vici VI.C.212.053).


\bibliography{refs}

\begin{thebibliography}{33}
\providecommand{\natexlab}[1]{#1}
\providecommand{\url}[1]{\texttt{#1}}
\expandafter\ifx\csname urlstyle\endcsname\relax
  \providecommand{\doi}[1]{doi: #1}\else
  \providecommand{\doi}{doi: \begingroup \urlstyle{rm}\Url}\fi

\bibitem[Bahdanau et~al.(2015)Bahdanau, Cho, and Bengio]{Bahdanau2014NeuralMT}
Bahdanau, D., Cho, K., and Bengio, Y.
\newblock Neural machine translation by jointly learning to align and translate.
\newblock In \emph{International Conference on Learning Representations ({ICLR})}, 2015.

\bibitem[Bal{\i}n et~al.(2019)Bal{\i}n, Abid, and Zou]{conc-autoencoder}
Bal{\i}n, M.~F., Abid, A., and Zou, J.
\newblock Concrete autoencoders: Differentiable feature selection and reconstruction.
\newblock In Chaudhuri, K. and Salakhutdinov, R. (eds.), \emph{Proceedings of the 36th International Conference on Machine Learning}, volume~97 of \emph{Proceedings of Machine Learning Research}, pp.\  444--453. PMLR, 09--15 Jun 2019.
\newblock URL \url{https://proceedings.mlr.press/v97/balin19a.html}.

\bibitem[Conwell \& Ullman(2022)Conwell and Ullman]{conwell2022testing}
Conwell, C. and Ullman, T.
\newblock Testing relational understanding in text-guided image generation, 2022.

\bibitem[Erhan et~al.(2013)Erhan, Goodfellow, Cukierski, and Bengio]{fer2013}
Erhan, D., Goodfellow, I., Cukierski, W., and Bengio, Y.
\newblock Challenges in representation learning: Facial expression recognition challenge, 2013.
\newblock URL \url{https://kaggle.com/competitions/challenges-inrepresentation-learning-facial-expression-recognition-challenge}.

\bibitem[Feng et~al.(2023)Feng, He, Fu, Jampani, Akula, Narayana, Basu, Wang, and Wang]{feng2023trainingfree}
Feng, W., He, X., Fu, T.-J., Jampani, V., Akula, A., Narayana, P., Basu, S., Wang, X.~E., and Wang, W.~Y.
\newblock Training-free structured diffusion guidance for compositional text-to-image synthesis, 2023.

\bibitem[Garnelo et~al.(2018{\natexlab{a}})Garnelo, Rosenbaum, Maddison, Ramalho, Saxton, Shanahan, Teh, Rezende, and Eslami]{pmlr-v80-garnelo18a}
Garnelo, M., Rosenbaum, D., Maddison, C., Ramalho, T., Saxton, D., Shanahan, M., Teh, Y.~W., Rezende, D.~J., and Eslami, S. M.~A.
\newblock Conditional neural processes.
\newblock In \emph{International Conference on Machine Learning ({ICML})}, volume~80 of \emph{Proceedings of Machine Learning Research}, pp.\  1690--1699, 2018{\natexlab{a}}.

\bibitem[Garnelo et~al.(2018{\natexlab{b}})Garnelo, Schwarz, Rosenbaum, Viola, Rezende, Eslami, and Teh]{DBLP:journals/corr/abs-1807-01622}
Garnelo, M., Schwarz, J., Rosenbaum, D., Viola, F., Rezende, D.~J., Eslami, S. M.~A., and Teh, Y.~W.
\newblock Neural processes.
\newblock \emph{CoRR}, abs/1807.01622, 2018{\natexlab{b}}.

\bibitem[Glorot \& Bengio(2010)Glorot and Bengio]{pmlr-v9-glorot10a}
Glorot, X. and Bengio, Y.
\newblock Understanding the difficulty of training deep feedforward neural networks.
\newblock In Teh, Y.~W. and Titterington, M. (eds.), \emph{Proceedings of the Thirteenth International Conference on Artificial Intelligence and Statistics}, volume~9 of \emph{Proceedings of Machine Learning Research}, pp.\  249--256, Chia Laguna Resort, Sardinia, Italy, 13--15 May 2010. PMLR.
\newblock URL \url{https://proceedings.mlr.press/v9/glorot10a.html}.

\bibitem[Gordon et~al.(2020)Gordon, Bruinsma, Foong, Requeima, Dubois, and Turner]{Gordon2020Convolutional}
Gordon, J., Bruinsma, W.~P., Foong, A. Y.~K., Requeima, J., Dubois, Y., and Turner, R.~E.
\newblock Convolutional conditional neural processes.
\newblock In \emph{International Conference on Learning Representations ({ICLR})}, 2020.

\bibitem[He et~al.(2022)He, Chen, Xie, Li, Doll{\'a}r, and Girshick]{he2022masked}
He, K., Chen, X., Xie, S., Li, Y., Doll{\'a}r, P., and Girshick, R.
\newblock Masked autoencoders are scalable vision learners.
\newblock In \emph{Proceedings of the IEEE/CVF Conference on Computer Vision and Pattern Recognition}, pp.\  16000--16009, 2022.

\bibitem[Jang et~al.(2017)Jang, Gu, and Poole]{jang2017categorical}
Jang, E., Gu, S., and Poole, B.
\newblock Categorical reparameterization with gumbel-softmax.
\newblock In \emph{International Conference on Learning Representations ({ICLR})}, 2017.

\bibitem[Jha et~al.(2022)Jha, Gong, Wang, Turner, and Yao]{arxiv.2209.00517}
Jha, S., Gong, D., Wang, X., Turner, R.~E., and Yao, L.
\newblock The neural process family: Survey, applications and perspectives, 2022.

\bibitem[Johnson et~al.(2017)Johnson, Hariharan, van~der Maaten, Fei-Fei, Lawrence~Zitnick, and Girshick]{johnson2017clevr}
Johnson, J., Hariharan, B., van~der Maaten, L., Fei-Fei, L., Lawrence~Zitnick, C., and Girshick, R.
\newblock {CLEVR}: A diagnostic dataset for compositional language and elementary visual reasoning.
\newblock In \emph{Proceedings of the IEEE Conference on Computer Vision and Pattern Recognition}, 2017.

\bibitem[Kawano et~al.(2021)Kawano, Kumagai, Sannai, Iwasawa, and Matsuo]{kawano2021group}
Kawano, M., Kumagai, W., Sannai, A., Iwasawa, Y., and Matsuo, Y.
\newblock Group equivariant conditional neural processes.
\newblock In \emph{International Conference on Learning Representations ({ICLR})}, 2021.

\bibitem[Kingma \& Welling(2014)Kingma and Welling]{kingma2014autoencoding}
Kingma, D.~P. and Welling, M.
\newblock Auto-encoding variational bayes.
\newblock In \emph{International Conference on Learning Representations ({ICLR})}, 2014.

\bibitem[Krizhevsky \& Hinton(2009)Krizhevsky and Hinton]{Krizhevsky_2009_17719}
Krizhevsky, A. and Hinton, G.
\newblock Learning multiple layers of features from tiny images.
\newblock Technical report, University of Toronto, 2009.

\bibitem[Liu et~al.(2015)Liu, Luo, Wang, and Tang]{liu2015faceattributes}
Liu, Z., Luo, P., Wang, X., and Tang, X.
\newblock Deep learning face attributes in the wild.
\newblock In \emph{International Conference on Computer Vision ({ICCV})}, pp.\  3730--3738, 2015.

\bibitem[Liu et~al.(2022)Liu, Mao, Wu, Feichtenhofer, Darrell, and Xie]{liu2022convnet}
Liu, Z., Mao, H., Wu, C.-Y., Feichtenhofer, C., Darrell, T., and Xie, S.
\newblock A convnet for the 2020s.
\newblock 2022.

\bibitem[Locatello et~al.(2020)Locatello, Weissenborn, Unterthiner, Mahendran, Heigold, Uszkoreit, Dosovitskiy, and Kipf]{locatello2020objectcentric}
Locatello, F., Weissenborn, D., Unterthiner, T., Mahendran, A., Heigold, G., Uszkoreit, J., Dosovitskiy, A., and Kipf, T.
\newblock Object-centric learning with slot attention, 2020.

\bibitem[Loshchilov \& Hutter(2019)Loshchilov and Hutter]{loshchilov2018decoupled}
Loshchilov, I. and Hutter, F.
\newblock Decoupled weight decay regularization.
\newblock In \emph{International Conference on Learning Representations ({ICLR})}, 2019.

\bibitem[Maddison et~al.(2017)Maddison, Mnih, and Teh]{maddison2017the}
Maddison, C.~J., Mnih, A., and Teh, Y.~W.
\newblock The concrete distribution: {A} continuous relaxation of discrete random variables.
\newblock In \emph{International Conference on Learning Representations ({ICLR})}, 2017.

\bibitem[Mentzer et~al.(2023)Mentzer, Minnen, Agustsson, and Tschannen]{mentzer2023finite}
Mentzer, F., Minnen, D., Agustsson, E., and Tschannen, M.
\newblock Finite scalar quantization: Vq-vae made simple, 2023.

\bibitem[Mnih et~al.(2014)Mnih, Heess, Graves, and Kavukcuoglu]{10.5555/2969033.2969073}
Mnih, V., Heess, N., Graves, A., and Kavukcuoglu, K.
\newblock Recurrent models of visual attention.
\newblock In \emph{Advances in Neural Information Processing Systems ({NeuRIPS})}, pp.\  2204--2212, 2014.

\bibitem[Mnmoustafa(2017)]{tiny-imagenet}
Mnmoustafa, M.~A.
\newblock Tiny imagenet, 2017.
\newblock URL \url{https://kaggle.com/competitions/tiny-imagenet}.

\bibitem[Nassar et~al.(2018)Nassar, Wang, and Tumer]{DBLP:journals/corr/abs-1812-05212}
Nassar, M., Wang, X., and Tumer, E.
\newblock Conditional graph neural processes: {A} functional autoencoder approach.
\newblock \emph{Third Workshop on Bayesian Deep Learning (NeurIPS 2018)}, abs/1812.05212, 2018.

\bibitem[Pathak et~al.(2016)Pathak, Kr{\"{a}}henb{\"{u}}hl, Donahue, Darrell, and Efros]{Pathak2016ContextEF}
Pathak, D., Kr{\"{a}}henb{\"{u}}hl, P., Donahue, J., Darrell, T., and Efros, A.~A.
\newblock Context encoders: Feature learning by inpainting.
\newblock In \emph{Conference on Computer Vision and Pattern Recognition ({CVPR})}, pp.\  2536--2544, 2016.

\bibitem[Ramesh et~al.(2021)Ramesh, Pavlov, Goh, Gray, Voss, Radford, Chen, and Sutskever]{pmlr-v139-ramesh21a}
Ramesh, A., Pavlov, M., Goh, G., Gray, S., Voss, C., Radford, A., Chen, M., and Sutskever, I.
\newblock Zero-shot text-to-image generation.
\newblock In \emph{International Conference on Machine Learning ({ICML})}, volume 139 of \emph{Proceedings of Machine Learning Research}, pp.\  8821--8831, 2021.

\bibitem[Reddi et~al.(2018)Reddi, Kale, and Kumar]{j.2018on}
Reddi, S.~J., Kale, S., and Kumar, S.
\newblock On the convergence of adam and beyond.
\newblock In \emph{International Conference on Learning Representations ({ICLR})}, 2018.

\bibitem[Shi et~al.(2022)Shi, Siddharth, Torr, and Kosiorek]{shi2022adversarial}
Shi, Y., Siddharth, N., Torr, P.~H., and Kosiorek, A.~R.
\newblock Adversarial masking for self-supervised learning.
\newblock In \emph{International Conference on Machine Learning ({ICML})}, 2022.

\bibitem[Trockman \& Kolter(2023)Trockman and Kolter]{trockman2023patches}
Trockman, A. and Kolter, J.~Z.
\newblock Patches are all you need?
\newblock \emph{Transactions on Machine Learning Research}, 2023.
\newblock ISSN 2835-8856.
\newblock URL \url{https://openreview.net/forum?id=rAnB7JSMXL}.
\newblock Featured Certification.

\bibitem[van~den Oord et~al.(2017)van~den Oord, Vinyals, and Kavukcuoglu]{10.5555/3295222.3295378}
van~den Oord, A., Vinyals, O., and Kavukcuoglu, K.
\newblock Neural discrete representation learning.
\newblock In \emph{Advances in Neural Information Processing Systems ({NeuRIPS})}, pp.\  6306--6315, 2017.

\bibitem[Xiao et~al.(2017)Xiao, Rasul, and Vollgraf]{xiao2017}
Xiao, H., Rasul, K., and Vollgraf, R.
\newblock Fashion-mnist: a novel image dataset for benchmarking machine learning algorithms, 2017.
\newblock URL \url{https://arxiv.org/abs/1708.07747}.

\bibitem[Zaheer et~al.(2017)Zaheer, Kottur, Ravanbakhsh, P{\'{o}}czos, Salakhutdinov, and Smola]{NIPS2017_f22e4747}
Zaheer, M., Kottur, S., Ravanbakhsh, S., P{\'{o}}czos, B., Salakhutdinov, R., and Smola, A.~J.
\newblock Deep sets.
\newblock In \emph{Advances in Neural Information Processing Systems ({NeuRIPS})}, pp.\  3391--3401, 2017.

\end{thebibliography}
\bibliographystyle{icml2024}

\newpage
\appendix
\onecolumn
\section{Implementation and Training of PPS-VAE}
\label{append:imp_and_train_details}
We parameterise the \ppsvae\ with CNN neural networks. More concretely, we use convolutional blocks similar to ConvNetXt \citep{liu2022convnet},  with \textit{Leaky ReLU} activation function. We found that with \textit{GELU} activation function training of PPS-VAE can be unstable. Also, we do not decrease the H\texttimes W dimensions of the original image, hence the induced $x_{1:M} \in \{0,1\}^{B \times H\times W\times 1}$ and $y_{1:M} \in [0,1]^{B\times H\times W\times C}$, where $B$ is the batch size. However, we represent the latent variable $a$ in a vector such that $a \in \mathbb{R}^{B\times D}$. We set $D$ to be 32, however any other values would work.

We optimise the parameters of the model with the AdamW \citep{loshchilov2018decoupled} optimiser, setting the learning rate to  $2*10^{-4}$ and we also enable the amsgrad \citep{j.2018on}. We train the PPS-VAE  for 200 epochs This is sufficient for the models to converge on the  datasets (\textbf{we provide code, which includes the implementation of the model}).

\subsection{Baseline Models}
\label{append:baselines}
\subsubsection{VQ-VAE}
The encoder of the VQ-VAE comprises of the 2 vanilla convolutional layers and 3 ResNet blocks. The decoder comprises of the 3 ResNet blocks and two transposed convolutions.  Between the layers we insert \textit{ReLU} activation function. The codebook is initialised with the \textit{xavier uniform} initialiser \cite{pmlr-v9-glorot10a}.  The latent representation of an image $z \in \mathbb{R}^{B\times J\times S\times D}$, where $J$ and $S$ are the reduced height and width of the original image and $D$ is the dimensionality of the vectors in the codebook. For each of the datasets  the number of the scalars in the codebook matches the number of elements in an original image. For example if an image has 3 colour channels and resolution of 64x64 then the total number of the scalar elements in the code book will be 64*64*3. The are multiple of ways to achieve this, we stick to the following. We set the number of vectors in the codebook to 64 and the dimensionality of the vectors to 64*3.  We optimise the parameters of the model with the AdamW optimiser, setting the learning rate to  $2*10^{-4}$ and we also enable the amsgrad. We train the models for 200 epochs.
\subsubsection{FSQ-VAE}
For FSQ-VAE we reuse the encoder and the decoder architecture of the VQ-VAE. We set the following number of of levels per
channel: for the colored images: $[8,8,7,6,5]$ and $[6,6,5,5,5]$ for black and white images. As in the VQ-VAE we choose the leves to roughly match the number of elements in an image. We optimise the parameters of the model with the AdamW optimiser, setting the learning rate to  $2*10^{-4}$ and we also enable the amsgrad. We train the models for 200 epochs.

\subsubsection{PPS-CAE}
 The encoder of PPS-CAE model comprises of $M$ 64*64 learnable parameters. These parameters are used to parameterise the Gumbel-Softmax distribution. We use the same $ p_\theta(\yt | \xt, \xm, \ym)$ decoder as for PPS-VAE.  We optimise the parameters of the model with the AdamW optimiser, setting the learning rate to  $2*10^{-4}$ and we also enable the amsgrad. We train the models for 200 epochs.

\section{Log Marginal Likelihood Estimation}
\label{append:log_marg_est}
In this section we show how we estimate log marginal likelihood of $\log p_\theta(\bm{y}|M)$. 
\subsection{Estimation for \ppsvae}
\[\log p_\theta(\bm{y}|M) \approx \log \left[\frac{1}{N}  \sum_{i=1}^{N}\frac{p_\theta(\bm{a}^i, \bm{x}^i, \bm{y}| M)}{q_\phi(\bm{a}^i, \xm^i|\bm{y}, M)} \right]; \qquad \bm{a}^i, \xm^i \sim q_\phi(\bm{a}, \xm|\bm{y}, M) \]
\subsection{Estimation for ConvCNP and PPS-CAE}
Given the CNP's generative model:
\[p_\theta(\bm{x}, \bm{y}| M)
  \!=\! p_\theta(\xm)\; p_\theta(\ym | \xm)\; p_\theta(\yt | \xt, \xm, \ym)\]
we estimate the log marginal likelihood as:
\[\log p_\theta(\bm{y}|M) \approx \log \left[\frac{1}{N} \sum_{i=1}^{N} p_\theta(\yt | \xt^i, \xm^i, \ym^i)\right]; \qquad \xm^i \sim p_\theta(\xm), \]
where $p_\theta(\ym | \xm)$ is delta function (=1), because of the deterministic look of $\ym$ values.

\section{Log Marginal Likelihood vs $M$}
\label{append:log_marginal_vs_M}

\begin{table}[H]
  \setlength{\tabcolsep}{3pt}
  \centering
  \caption{Estimated $\log p_\theta(\bm{y}|M) (\uparrow)$ for \ppsvae\ (see Appendix \ref{append:log_marg_est})  with 800 samples.}
  \label{table:}
  \begin{tabular}{@{}lcccc@{}}
    \toprule
    &  FER2013 &  CelA & CLEVR & t-ImageNet \\
    \midrule
    $M=32$ &  4111 &  11569 &  16529 &  15645 \\
    $M=64$ &  4711 &  13251 &  16604 &  16269 \\
    $M=128$ &  \cellcolor[HTML]{a8ddb5} 4951 &  \cellcolor[HTML]{a8ddb5}14210 &  \cellcolor[HTML]{a8ddb5}16611&  \cellcolor[HTML]{a8ddb5}16324\\
    \bottomrule
  \end{tabular}
\end{table}
\raggedbottom

\section{Inductive Bias: MLP CNP}
\label{append:mlp_ind_bias}

\begin{figure}[H]

\subcaptionbox{$M=30$\label{figure: pps 30 (fashionmnist)}}{\includegraphics[width=\textwidth, height=2.7cm]{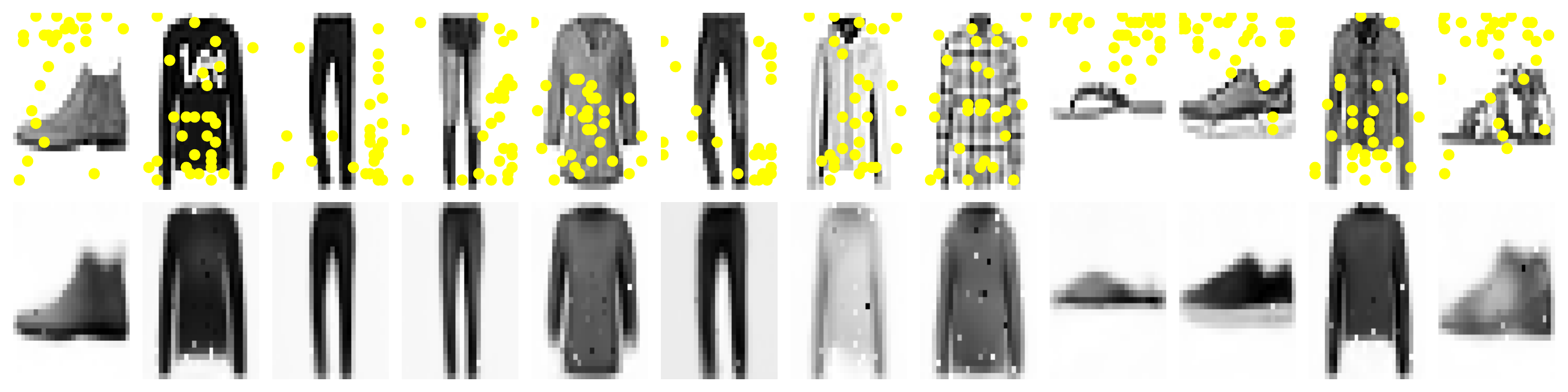}}
\caption{Visualisation of the spatial arrangement of points in the context sets for the CNP decoder parameterised by the MLP --- conducted on the FashionMNIST dataset. The first row corresponds to the original image, together with the inferred context set denoted by the yellow circles. The second row corresponds to the reconstructed images. The context sets inferred on the test images.}
\end{figure}

In the earlier version of the PPS-VAE model, we found that the parametisation of the model with MLP layers as in \cite{pmlr-v80-garnelo18a} may bias the model to infer points around the edges of an image (see Figure \ref{figure: pps 30 (fashionmnist)}). 

\section{Parallel vs Autoregressive Encoder}
\label{append:parallel_vs_ind_enc}
\begin{table*}[h!]
  \setlength{\tabcolsep}{5pt}
  \centering
  \caption{Object classification for two datasets: FashionMNIST \citep{xiao2017} and CIFAR10 \citep{Krizhevsky_2009_17719}. $M=60$. Resnet-18 classifiers trained from scratch over three seeds with early stopping, reporting mean F1-macro scores. \ppsvaea\ is the \ppsvae\ with the autoregressive encoder used in the main paper. \ppsvaei\ is the \ppsvae\ with independent encoder over $\xm$: \(q_\phi(\xm | \bm{y}) = \prod_{m=1}^M \mathop{GS}(x_m | h_\phi(y))\), where $h_\phi$ is a parametrised
neural network that transform inputs to distribution parameters }
  \label{table:independent_vs_autoreg_enc}
  \scalebox{0.83}{%
  \begin{tabular}{@{}l@{\quad}lcc@{}}
    \toprule
    & & \bf FashionMNIST & \bf CIFAR-10  \\
    \midrule
    & \ppsvaea\ (points) &  88.0 $\pm$ 0.0  &  76.7 $\pm$ 0.5   \\
    & \ppsvaei\ (points) &  86.0 $\pm$ 0.0  &  68.0 $\pm$ 0.0   \\
    \bottomrule
  \end{tabular}}
\end{table*}

\section{Classification Results: Number of Points in Context Sets vs Classification Performance}
\label{append:class_perfm_vs_M}

\begin{table*}[h!]
  \setlength{\tabcolsep}{4pt}
  \centering
  \caption{Object classification. \ppsvae\ ($M$ vs F1). Classifiers trained over three seeds with early stopping, reporting mean F1-macro scores. A:13 --- Chubby, A:20 --- Male, A:25 --- Oval Face.}
  \label{table:}
  \scalebox{1.}{%
  \begin{tabular}{@{}l@{\quad}lcccccc@{}}
    \toprule
    & &   CelA (A:13) &  CelA (A:20) &  CelA (A:25)  & FER2013 &  CLEVR & t-ImageNet  \\
    \midrule
   
    & $M=32$ & 57.62 $\pm$ 0.85 &  88.28 $\pm$ 0.07  & 57.57 $\pm$ 0.88 &   29.19 $\pm$ 3.73 &  68.58 $\pm$ 0.17  &  13.59 $\pm$ 1.72    \\
     & $M=64$ & 63.46 $\pm$ 0.66 &  91.81 $\pm$ 0.03  & 60.04 $\pm$ 0.70 &   40.18 $\pm$ 0.41 &  84.91 $\pm$ 1.43  &  21.49 $\pm$ 1.75    \\
    & $M=128$& 69.00 $\pm$ 0.38 &  94.86 $\pm$ 0.12 & 62.13 $\pm$ 0.50  &  46.72 $\pm$ 0.62  &  90.21 $\pm$ 0.28 &  29.56 $\pm$ 0.27    \\
    \bottomrule
  \end{tabular}}
\end{table*}

\section{Classification Results: Increasing Number of Points in Context Sets at Inference Time}
\label{append:inc_M_capacity}
\begin{table}[H]
  \setlength{\tabcolsep}{4pt}
  \centering
  \caption{Object classification. \ppsvae\ ($M$ vs F1). Classifiers trained over three seeds with early stopping, reporting mean F1-macro scores. A:13 --- Chubby, A:20 --- Male, A:25 --- Oval Face.}
  \label{table:}
  \scalebox{1.}{%
  \begin{tabular}{@{}lcccccc@{}}
    \toprule
    &   CelA (A:13) &  CelA (A:20) &  CelA (A:25)  & FER2013 &  CLEVR & t-ImageNet  \\
    \midrule
   
    $M=32 \rightarrow 64$ & 62.81 $\pm$ 0.50 &  91.34 $\pm$ 0.13  & 58.74 $\pm$ 0.70 &   39.57 $\pm$ 0.21 &  82.61 $\pm$ 1.65  &  20.30 $\pm$ 1.23    \\
    $M=32 \rightarrow 128$  & 65.54 $\pm$ 0.32 &  92.80 $\pm$ 0.11  & 60.26 $\pm$ 0.99 &   45.16 $\pm$ 0.33 &  87.59 $\pm$ 0.32 &  25.78 $\pm$ 0.20    \\[1ex]
    $M=64 \rightarrow 128$& 67.50 $\pm$ 0.68 &  94.11 $\pm$ 0.06 & 61.68 $\pm$ 0.42  &  47.23 $\pm$ 0.32  &  91.15 $\pm$ 0.37 &  27.98 $\pm$ 0.70    \\[1ex]
    $M=128\rightarrow 256$& 69.94 $\pm$ 0.50 &  95.70 $\pm$ 0.07 & 62.02 $\pm$ 0.50  &  51.61 $\pm$ 0.57  &  93.38 $\pm$ 0.64 &  33.93 $\pm$ 0.16  \\
    \bottomrule
  \end{tabular}}
\end{table}

\section{Classification Results: Evaluating Latent Variable $\bm{a}$}
\label{append:z_vs_a}
\begin{table}[H]
  \setlength{\tabcolsep}{4pt}
  \centering
  \caption{Object classification. Benchmarking latent variable $\bm{a}$ against vanilla VAE. Classifiers trained over three seeds with early stopping, reporting mean F1-macro scores. A:13 --- Chubby, A:20 --- Male, A:25 --- Oval Face.}
  \label{table:}
  \scalebox{1.}{%
  \begin{tabular}{@{}l@{\quad}lcccccc@{}}
    \toprule
    & &   CelA (A:13) &  CelA (A:20) &  CelA (A:25)  & FER2013 &  CLEVR & t-ImageNet  \\
    \midrule
   
    & VAE & 59.04 $\pm$ 0.66 &  86.26 $\pm$ 0.12  & 58.60 $\pm$ 0.27 &   36.06 $\pm$ 0.34 &  41.88 $\pm$ 0.28  &  10.05 $\pm$ 0.05    \\
    & \ppsvae\ ($\bm{a}$) & 54.26 $\pm$ 0.42 &  83.52 $\pm$ 0.09  & 55.42 $\pm$ 0.24 &   20.83 $\pm$ 0.13 &  39.62 $\pm$ 0.11  &  8.50 $\pm$ 0.13    \\
    \bottomrule
  \end{tabular}}
\end{table}
\paragraph{VAE Model:}
The encoder of the VAE baseline comprises of five convolutional layers: 3 are the vanilla convolutions with \textit{Leaky ReLU} activation function inserted between them and 2 vanilla convolutions to model the mean and variance of the variational posterior distribution, which is Gaussian. Both the mean and the variance are 32 dimensional vectors.  The architecture of the encoder resembles the parametrisation of $q_\phi(\bm{a} | \xm, \ym)$. The decoder comprises of five transposed convolutions with \textit{Leaky ReLU} activation function inserted between them.  We optimise the parameters of the model with the AdamW optimiser, setting the learning rate to  $2*10^{-4}$ and we also enable the amsgrad.

\section{Compute}
We run each experiment using the hardware specified in Table \ref{table:hardware}.
\begin{table}[H]
\setlength{\tabcolsep}{2pt}
\caption{Computing infrastructure.}
\centering
\scalebox{0.70}{
\begin{tabular}{l r  }
\toprule
  hardware & specification \\
 \midrule
    CPU  & AMD\textsuperscript{\textregistered} EPYC 7413 24-Core Processor \\
  GPU  & NVIDIA\textsuperscript{\textregistered} A40 x 1 \\
 \bottomrule
\end{tabular}
}
\label{table:hardware}
\end{table}
\section{Parameters Count}

\begin{table}[H]
  \setlength{\tabcolsep}{3pt}
  \centering
  \caption{Number of parameters in a model.}
  \label{table:num_param}
  \begin{tabular}{@{}lcccc@{}}
    \toprule
     & \ppsvae & FSQ-VAE & VQ-VAE & PPS-CAE  \\
    \midrule
    \# parameters &  6,183,740 &  10,541,832 &  11,511,747 & 5,278,982  \\
   
    \bottomrule
  \end{tabular}
\end{table}
To calculate total number of parameters in the model we use:

\begin{lstlisting}[language=Python]
params = sum(p.numel() for p in model.parameters() if p.requires_grad)
\end{lstlisting}

\section{Algorithm}
\begin{algorithm}[H]
   \caption{PPS-VAE}
   \label{alg: pps-vae-recurr-post}
\begin{algorithmic}
   \STATE // ** Inference **
   \STATE {\bfseries Input:}  $y \in \mathbb{R}^{C\times H\times W}$
   \STATE Initialize  $x_0 \in \{0,1\}^{C\times H\times W} = 0$,
    $x_{1:M} = [x_0]$
   \FOR{$i=1$ {\bfseries to} $M$}
    \STATE $x_i \sim q_{\phi}(x_i|y, x_{1:M}[0:i])$
    \STATE $x_{1:M}.append(x_i)$
   \ENDFOR
   \STATE $x_{1:M} = \text{sum}(x_{1:M}, \text{axis}=0) \in \{0,1\}^{1\times H\times W}$
   \STATE // points can be sampled twice, remove duplicates
   \STATE  $x_{1:M} =  x_{1:M} / x_{1:M}$
   \STATE $y_{1:M} = y * x_{1:M}$
   \STATE $a \sim q_\phi(a|x_{1:M},y_{1:M})$
   \STATE // ** Scoring **
   \STATE $D_{KL}$ = $(\log q_\phi(x_{1:M}|y)-\log p_\theta(x_{1:M}|a))$ + \\          
    \hspace{40pt} +$(q_\phi(a|y_{1:M},x_{1:M}) -p_{\theta}(a))$
   \STATE // get the target variables
   \STATE $x_{1:T} = 1 - x_{1:M}$, $y_{1:T} = y * x_{1:T}$
   \STATE loss($y_{1:M}$) = $\log p_\theta(y_{1:M}|x_{1:M},a)$
   \STATE loss($y_{1:T}$) = $\log p_\theta(y_{1:T}| y_{1:M}, x_{1:M};x_{1:T})$
\end{algorithmic}
\end{algorithm}

\section{Visualisation of Reconstructed Images}
\label{append:vis_rec_imgs}
\subsection{\ppsvae}
\subsubsection{Dataset: t-ImageNet}
\begin{figure}[H]
\centering
\subcaptionbox{$M=32$}{\includegraphics[width=\linewidth, height=3.1cm]{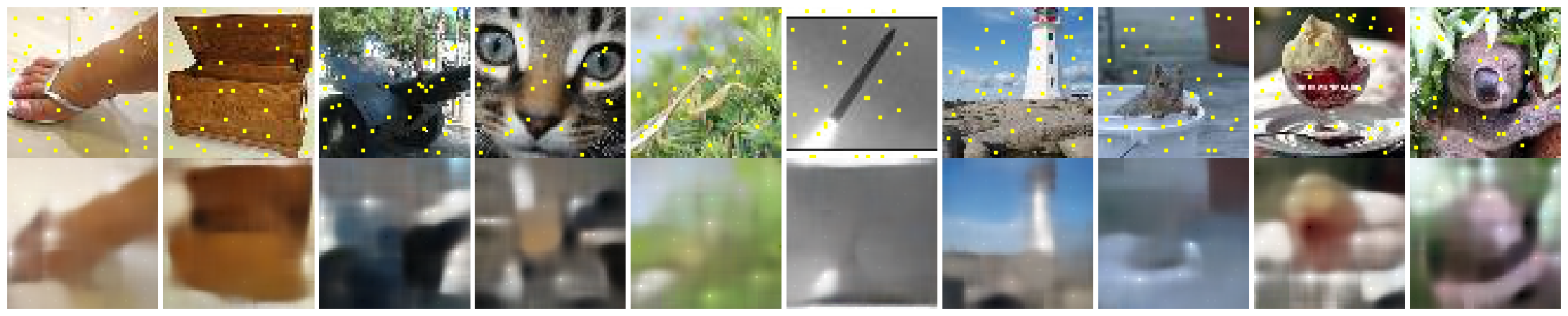}}
\subcaptionbox{$M=64$}{\includegraphics[ width=\linewidth, height=3.1cm]{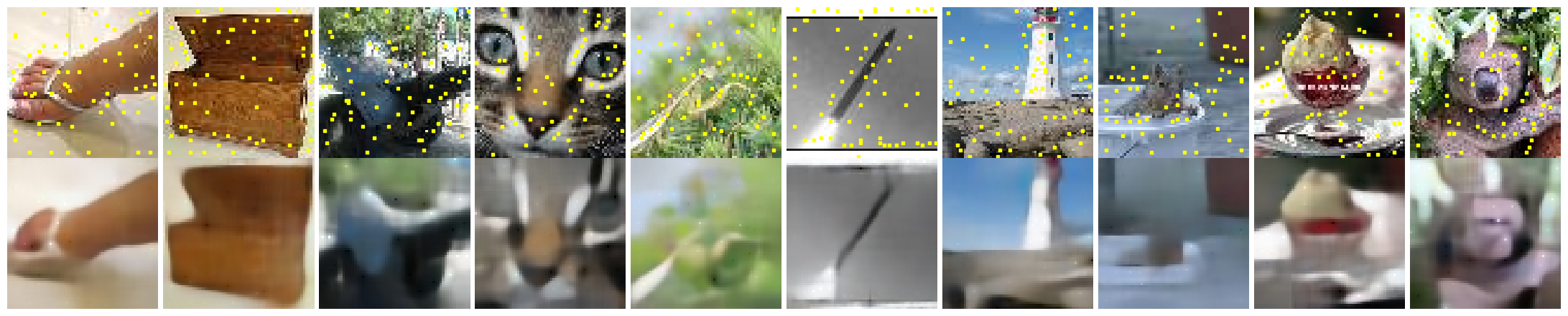}}
\subcaptionbox{$M=128$}{\includegraphics[width=\linewidth, height=3.1cm]{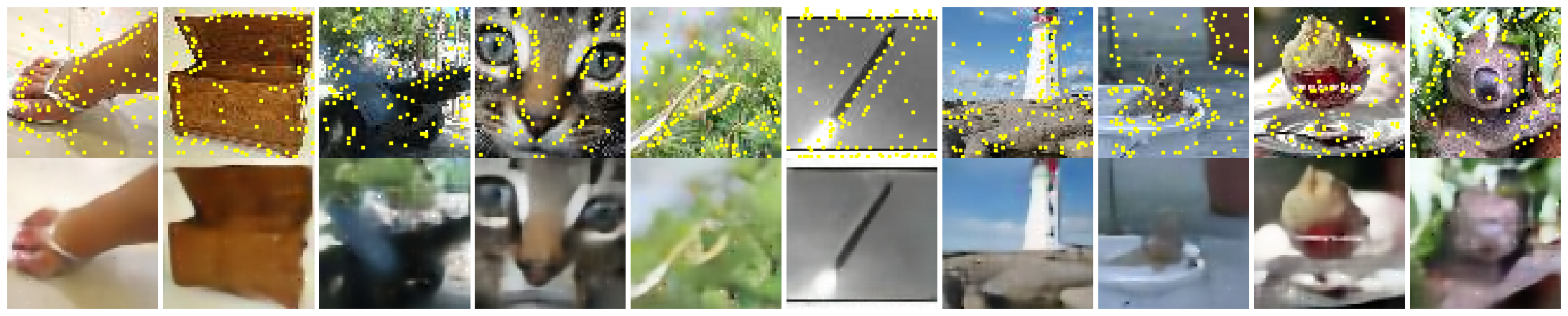}}
\caption{Visualisation of the spatial arrangement of points in the context sets. The first row corresponds to the original image, together with the inferred context set denoted by the yellow circles. The second row corresponds to the reconstructed images. The context sets inferred on the test images.}
\label{figure: }
\end{figure}

\subsubsection{Dataset: CLEVR}
\begin{figure}[H]
\centering
\subcaptionbox{$M=32$}{\includegraphics[width=\linewidth, height=3.1cm]{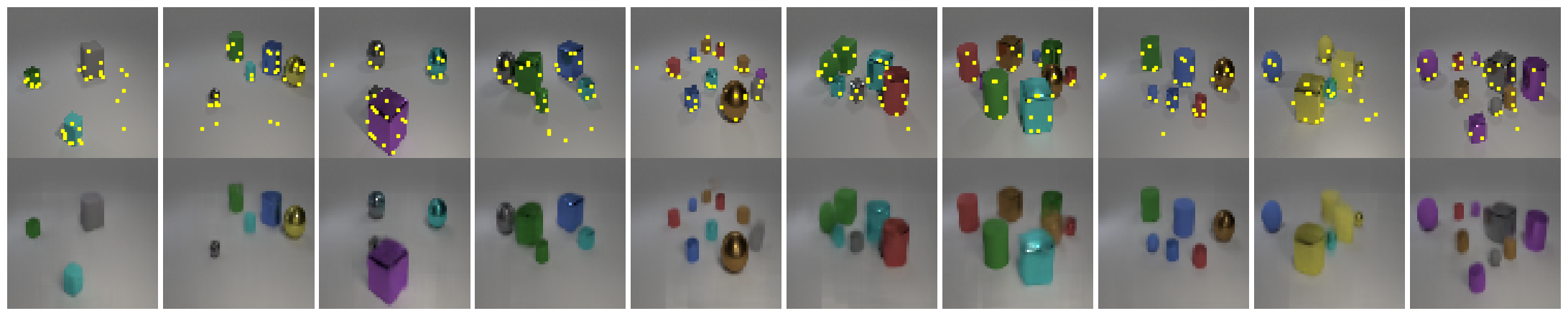}}
\subcaptionbox{$M=64$}{\includegraphics[ width=\linewidth, height=3.1cm]{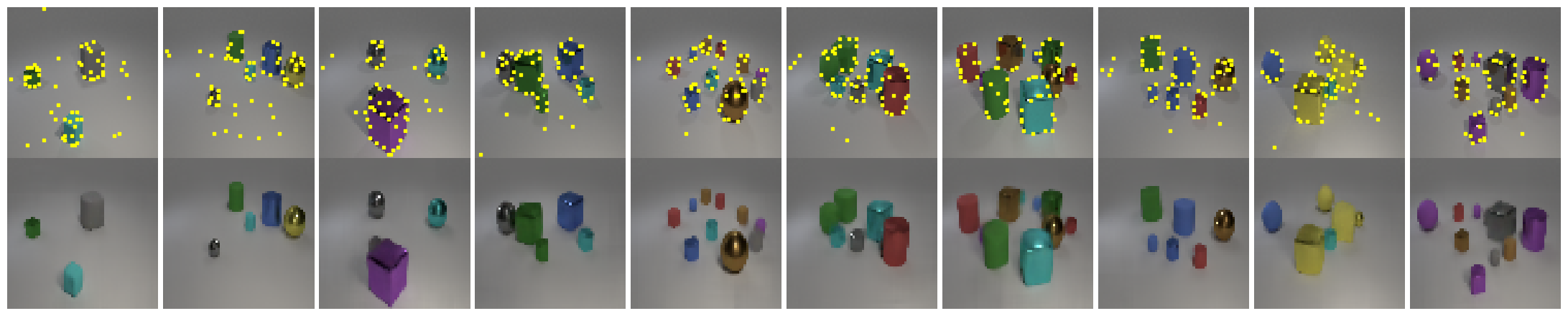}}
\subcaptionbox{$M=128$}{\includegraphics[width=\linewidth, height=3.1cm]{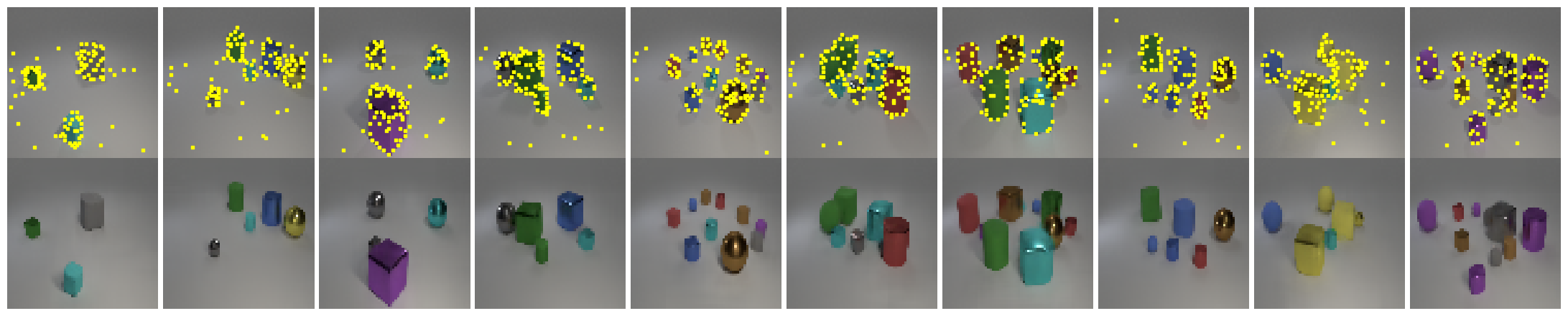}}
\caption{ Visualisation of the spatial arrangement of points in the context sets. The first row corresponds to the original image, together with the inferred context set denoted by the yellow circles. The second row corresponds to the reconstructed images. The context sets inferred on the test images.}
\label{figure: }
\end{figure}
\subsubsection{Dataset: CelebA}
\begin{figure}[H]
\centering
\subcaptionbox{$M=32$}{\includegraphics[width=\linewidth, height=3.1cm]{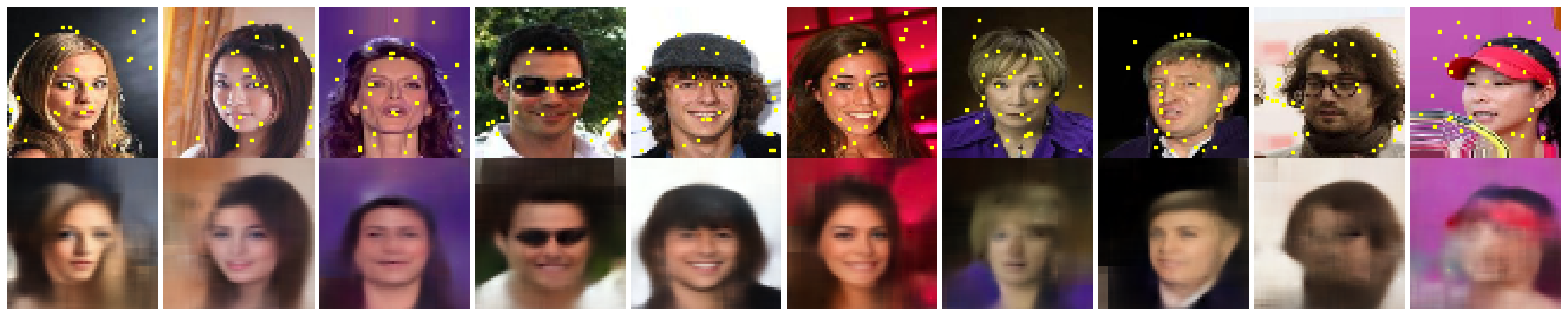}}
\subcaptionbox{$M=64$}{\includegraphics[ width=\linewidth, height=3.1cm]{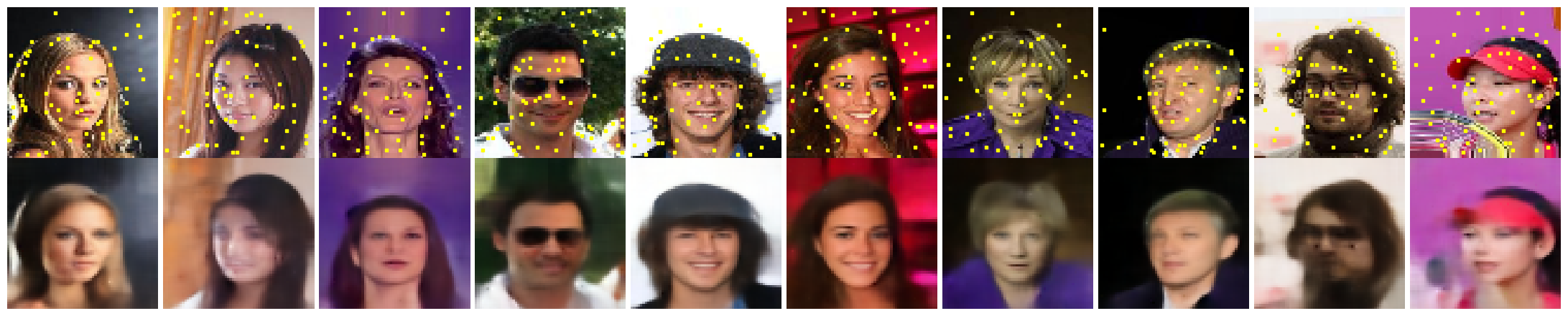}}
\subcaptionbox{$M=128$}{\includegraphics[width=\linewidth, height=3.1cm]{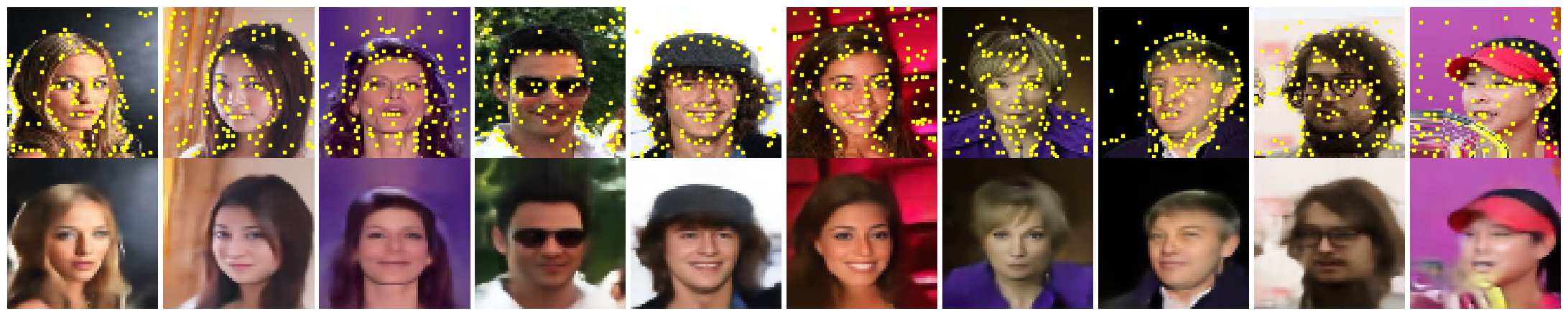}}
\caption{Visualisation of the spatial arrangement of points in the context sets. The first row corresponds to the original image, together with the inferred context set denoted by the yellow circles. The second row corresponds to the reconstructed images. The context sets inferred on the test images.}
\label{figure: }
\end{figure}
\subsubsection{Dataset: FER2013}
\begin{figure}[H]
\centering
\subcaptionbox{$M=32$}{\includegraphics[width=\linewidth, height=3.1cm]{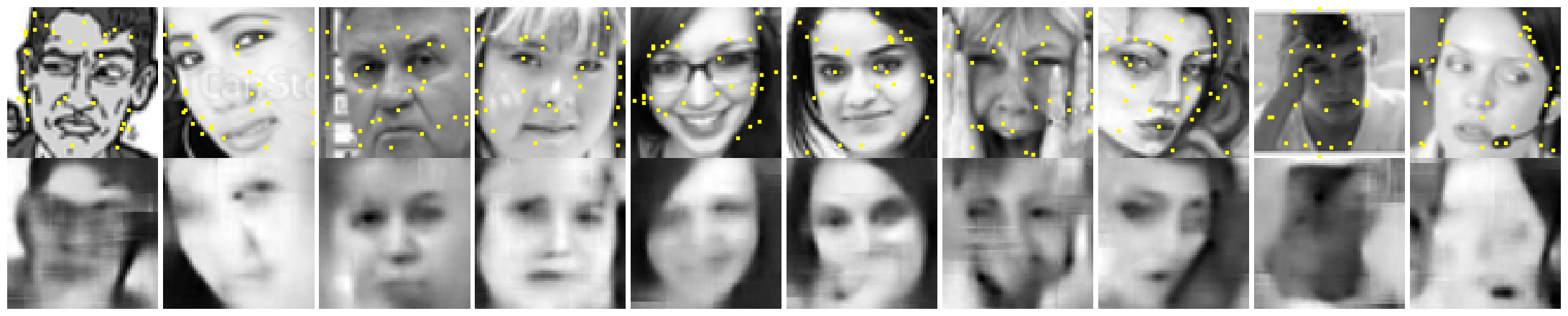}}
\subcaptionbox{$M=64$}{\includegraphics[ width=\linewidth, height=3.1cm]{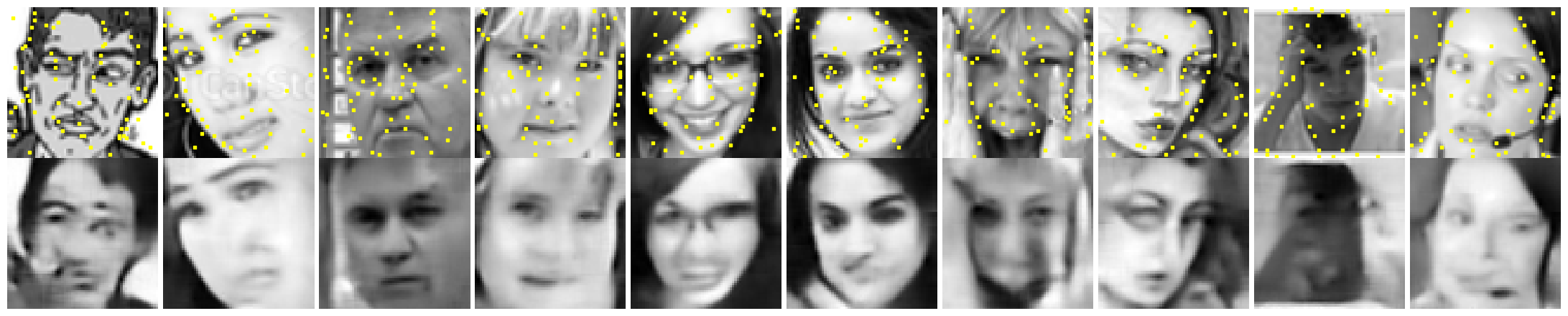}}
\subcaptionbox{$M=128$}{\includegraphics[width=\linewidth, height=3.1cm]{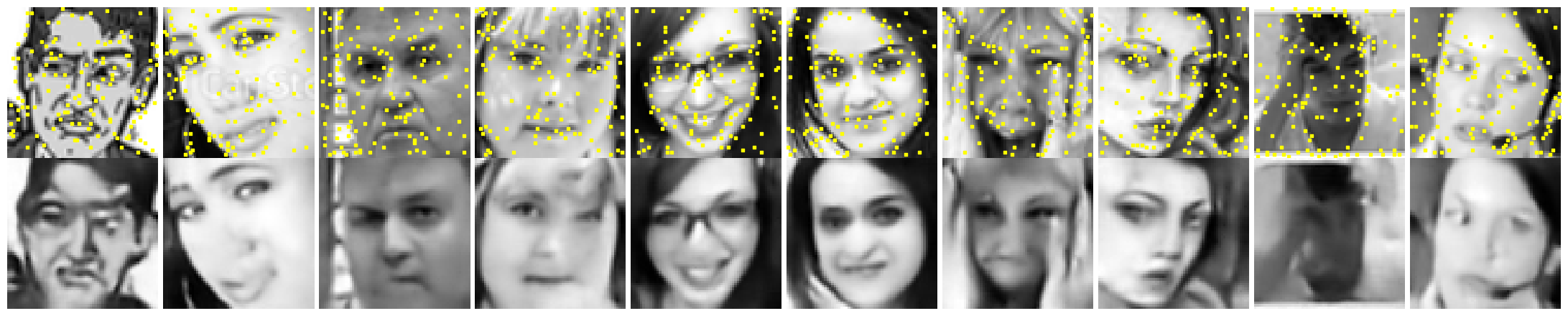}}
\caption{Visualisation of the spatial arrangement of points in the context sets. The first row corresponds to the original image, together with the inferred context set denoted by the yellow circles. The second row corresponds to the reconstructed images. The context sets inferred on the test images.}
\label{figure: }
\end{figure}
\subsection{VQ-VAE}
\subsubsection{Dataset: t-ImageNet}
\begin{figure}[H]
\centering
\includegraphics[width=\linewidth, height=3.1cm]{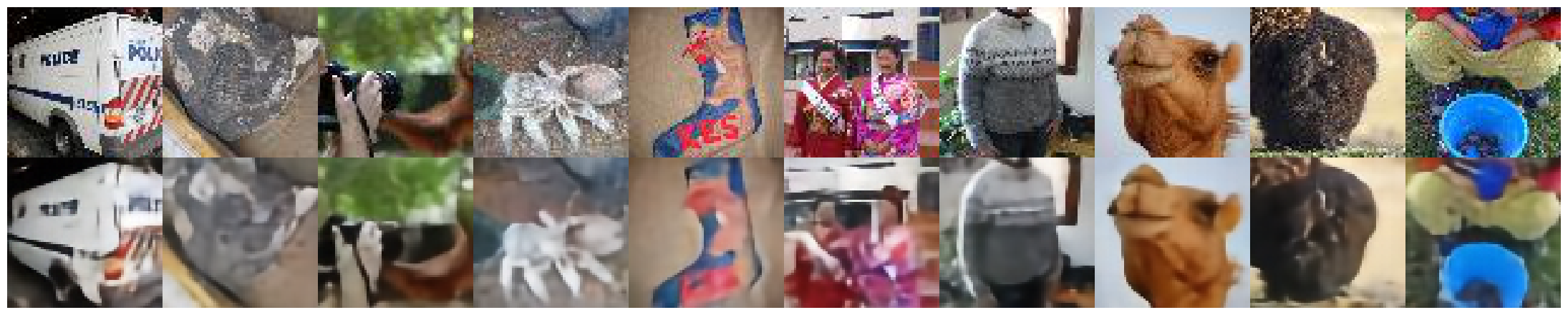}
\caption{The first row corresponds to the original image. The second row corresponds to the reconstructed images. Evaluated on the test images.}
\label{figure: }
\end{figure}
\subsubsection{Dataset: CLEVR}
\begin{figure}[H]
\centering
\includegraphics[width=\linewidth, height=3.1cm]{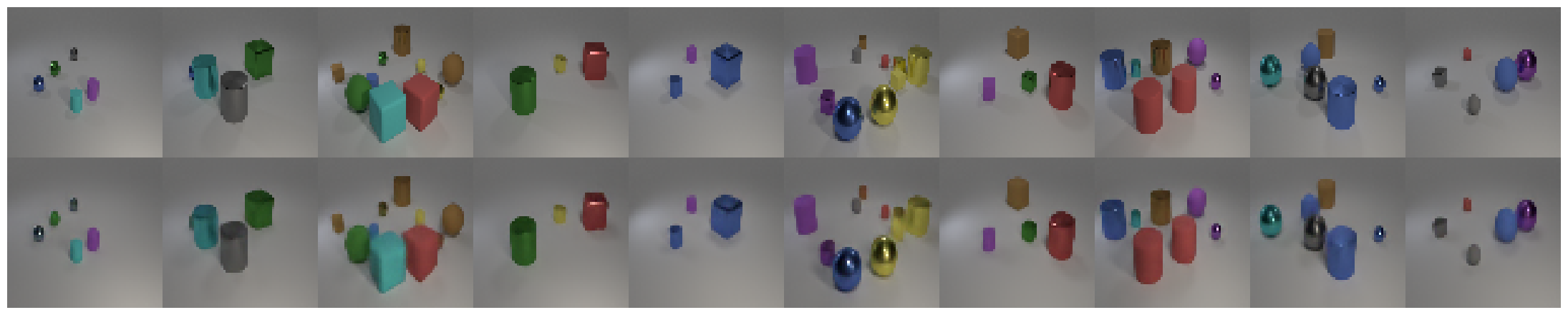}
\caption{The first row corresponds to the original image. The second row corresponds to the reconstructed images. Evaluated on the test images.}
\label{figure: }
\end{figure}
\subsubsection{Dataset: CelebA}
\begin{figure}[H]
\centering
\includegraphics[width=\linewidth, height=3.1cm]{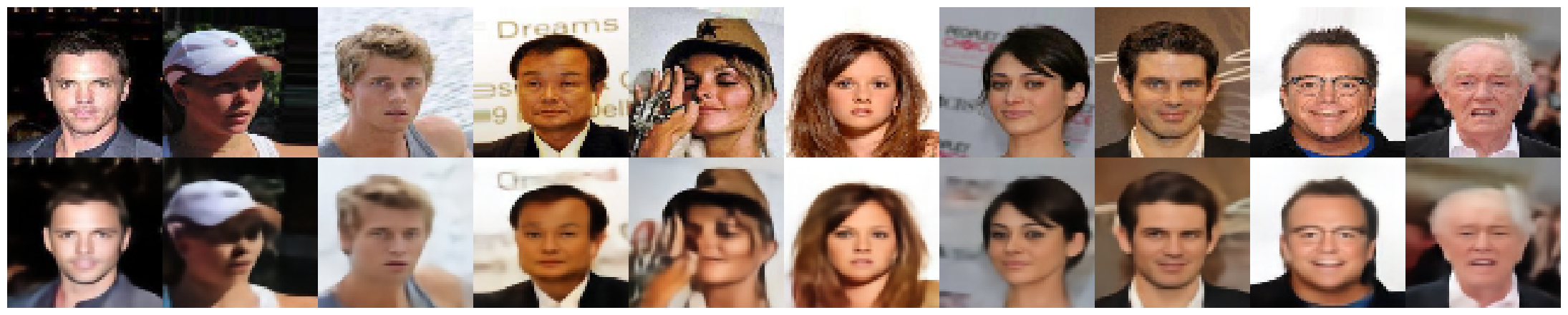}
\caption{The first row corresponds to the original image. The second row corresponds to the reconstructed images. Evaluated on the test images.}
\label{figure: }
\end{figure}
\subsubsection{Dataset: FER2013}
\begin{figure}[H]
\centering
\includegraphics[width=\linewidth, height=3.1cm]{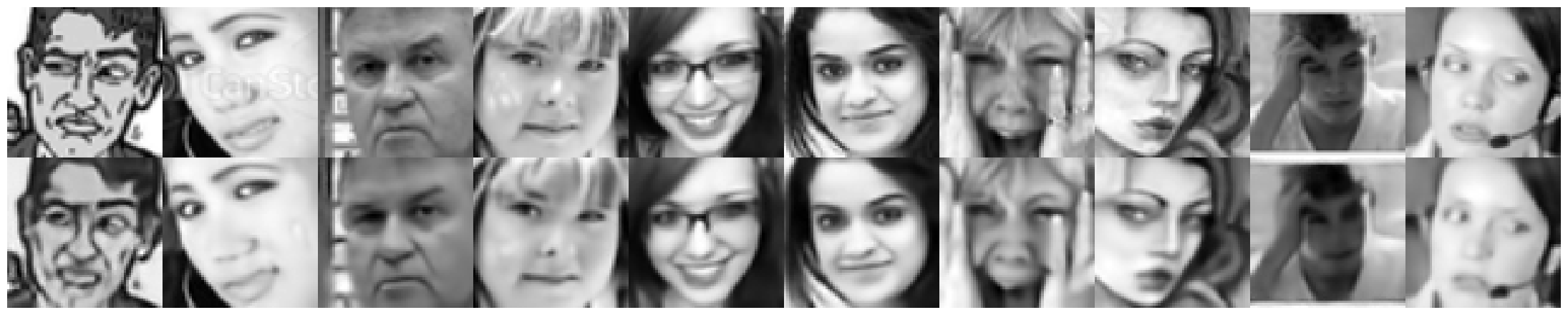}
\caption{The first row corresponds to the original image. The second row corresponds to the reconstructed images. Evaluated on the test images.}
\label{figure: }
\end{figure}

\subsection{FSQ-VAE}
\subsubsection{Dataset: t-ImageNet}
\begin{figure}[H]
\centering
\includegraphics[width=\linewidth, height=3.1cm]{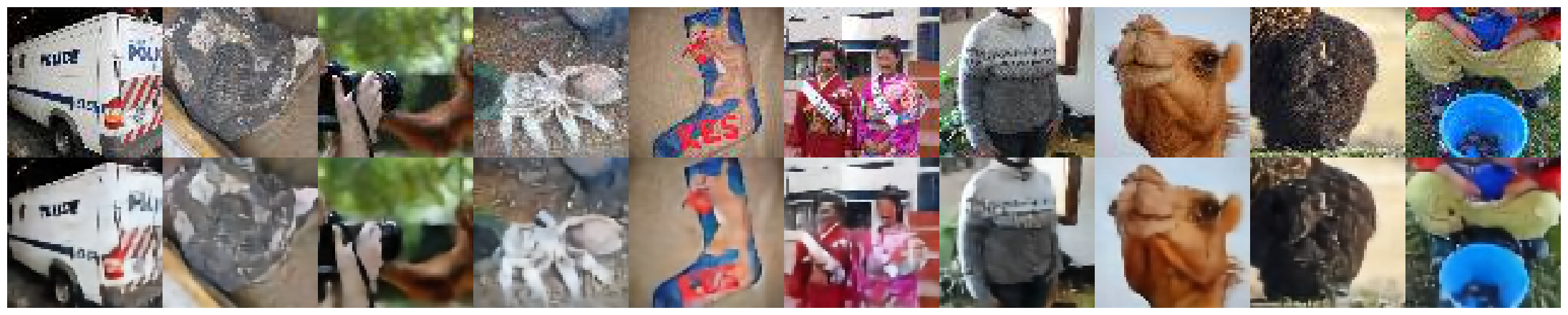}
\caption{The first row corresponds to the original image. The second row corresponds to the reconstructed images. Evaluated on the test images.}
\label{figure: }
\end{figure}
\subsubsection{Dataset: CLEVR}
\begin{figure}[H]
\centering
\includegraphics[width=\linewidth, height=3.1cm]{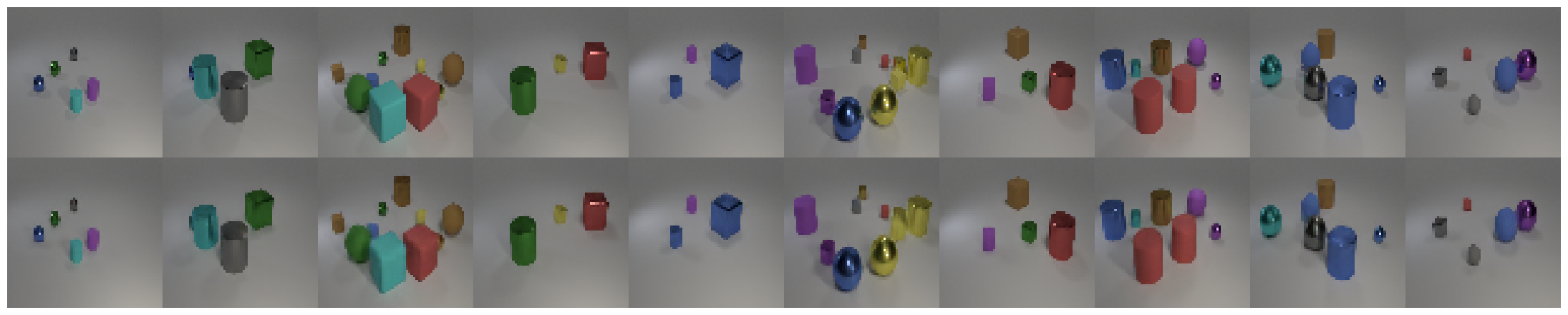}
\caption{The first row corresponds to the original image. The second row corresponds to the reconstructed images. Evaluated on the test images.}
\label{figure: }
\end{figure}
\subsubsection{Dataset: CelebA}
\begin{figure}[H]
\centering
\includegraphics[width=\linewidth, height=3.1cm]{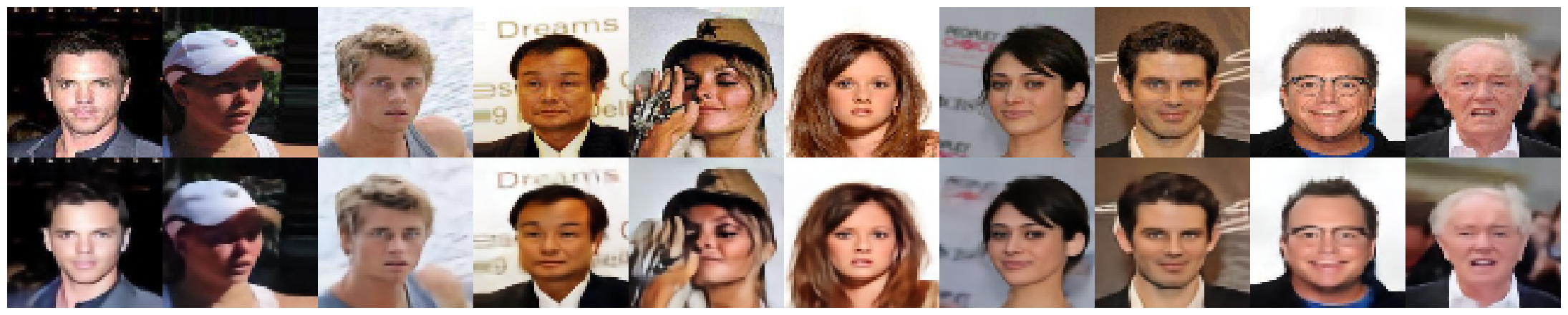}
\caption{The first row corresponds to the original image. The second row corresponds to the reconstructed images. Evaluated on the test images.}
\label{figure: }
\end{figure}
\subsubsection{Dataset: FER2013}
\begin{figure}[H]
\centering
\includegraphics[width=\linewidth, height=3.1cm]{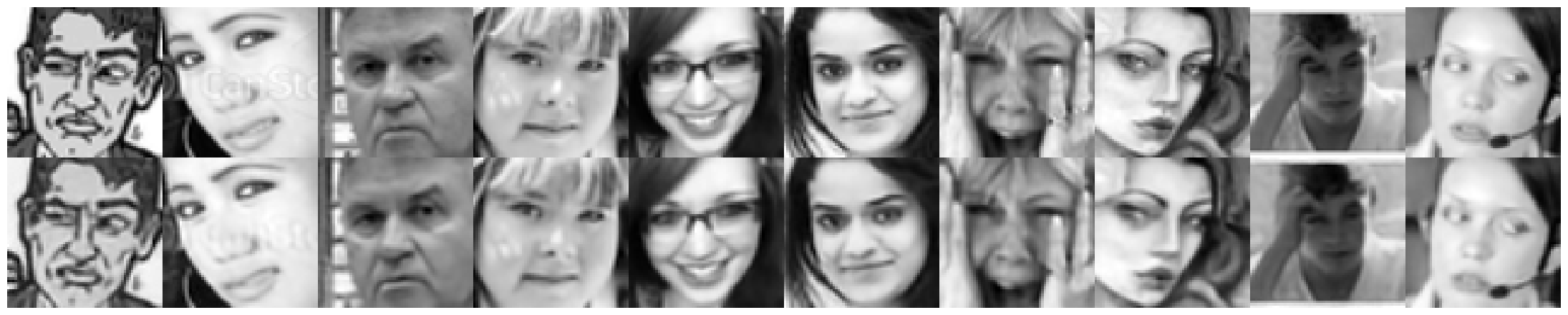}
\caption{The first row corresponds to the original image. The second row corresponds to the reconstructed images. Evaluated on the test images.}
\label{figure: }
\end{figure}

\section{Visualisation of Out-of-Distribution Reconstruction}
\label{append:vis_out_dist_rec}
\subsection{PPS-VAE}
\subsubsection{Training Dataset: t-ImageNet}
\begin{figure}[H]
\centering
\includegraphics[width=\linewidth, height=3.1cm]{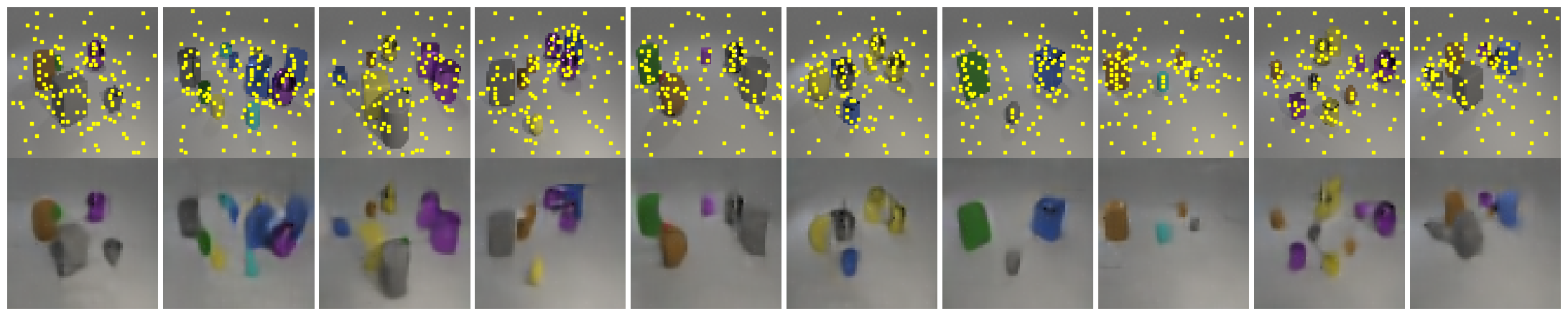}
\caption{Test dataset CLEVR. Visualisation of the spatial arrangement of points in the context sets. The first row corresponds to the original image, together with the inferred context set denoted by the yellow circles. The second row corresponds to the reconstructed images. The context sets inferred on the test images. }
\label{figure: }
\end{figure}
\begin{figure}[H]
\centering
\includegraphics[width=\linewidth, height=3.1cm]{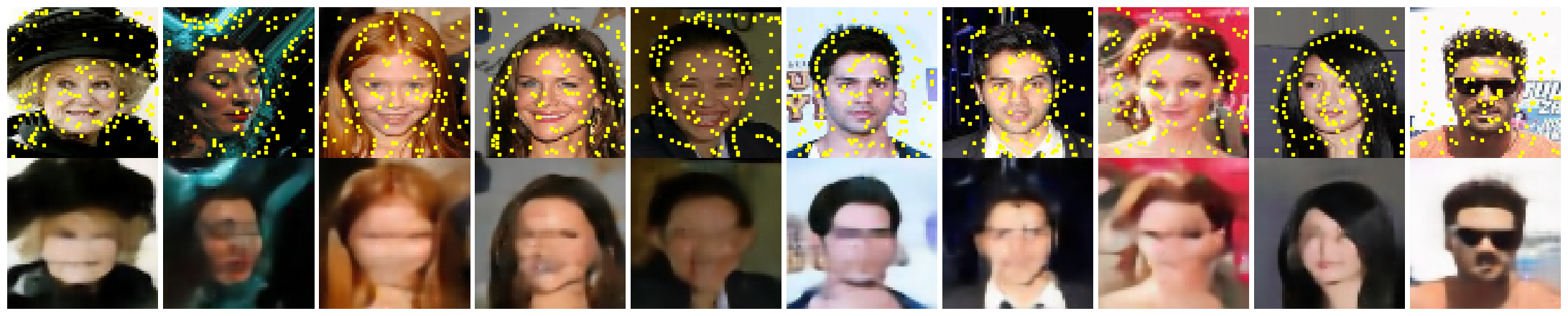}
\caption{Test dataset CelA. Visualisation of the spatial arrangement of points in the context sets. The first row corresponds to the original image, together with the inferred context set denoted by the yellow circles. The second row corresponds to the reconstructed images. The context sets inferred on the test images.}
\label{figure: }
\end{figure}
\subsubsection{Training Dataset: CLEVR}
\begin{figure}[H]
\centering
\includegraphics[width=\linewidth, height=3.1cm]{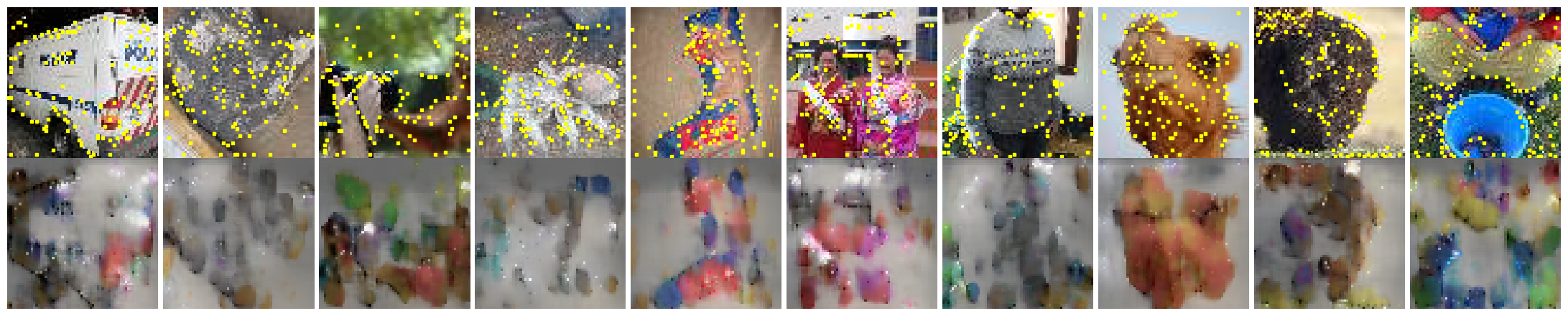}
\caption{Test dataset t-ImageNet. Visualisation of the spatial arrangement of points in the context sets. The first row corresponds to the original image, together with the inferred context set denoted by the yellow circles. The second row corresponds to the reconstructed images. The context sets inferred on the test images.}
\label{figure: }
\end{figure}
\begin{figure}[H]
\centering
\includegraphics[width=\linewidth, height=3.1cm]{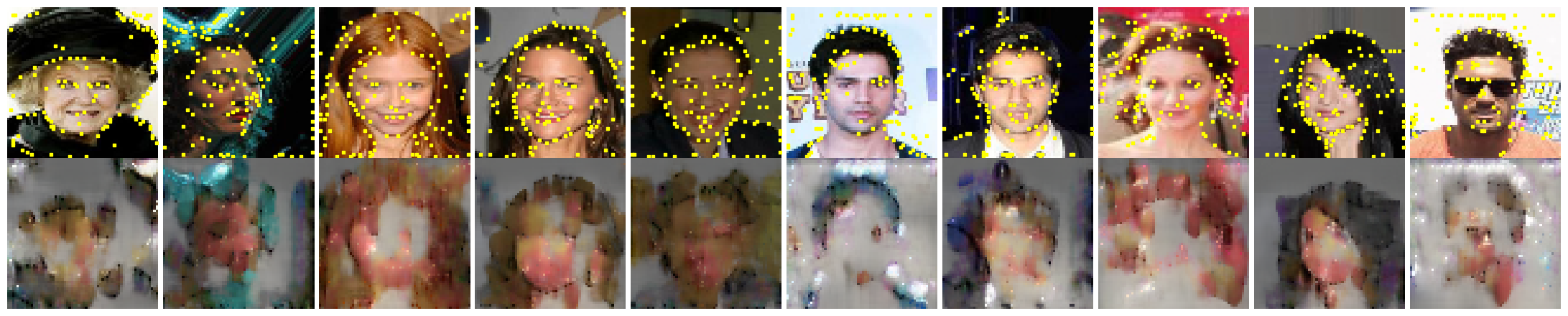}
\caption{Test dataset CelA. Visualisation of the spatial arrangement of points in the context sets. The first row corresponds to the original image, together with the inferred context set denoted by the yellow circles. The second row corresponds to the reconstructed images. The context sets inferred on the test images.}
\label{figure: }
\end{figure}
\subsubsection{Training Dataset: CelebA}
\begin{figure}[H]
\centering
\includegraphics[width=\linewidth, height=3.1cm]{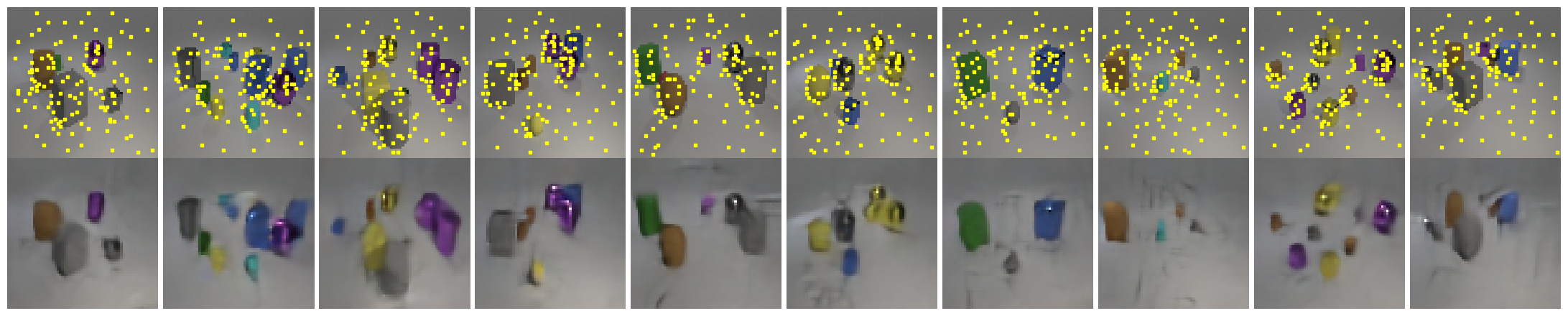}
\caption{Test dataset CLEVR. Visualisation of the spatial arrangement of points in the context sets. The first row corresponds to the original image, together with the inferred context set denoted by the yellow circles. The second row corresponds to the reconstructed images. The context sets inferred on the test images.}
\label{figure: }
\end{figure}
\begin{figure}[H]
\centering
\includegraphics[width=\linewidth, height=3.1cm]{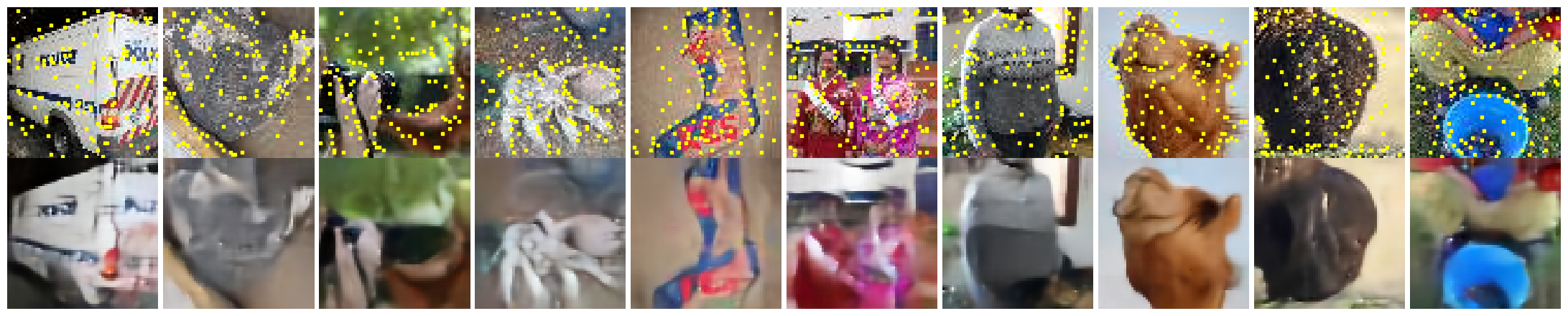}
\caption{Test dataset t-ImageNet. Visualisation of the spatial arrangement of points in the context sets. The first row corresponds to the original image, together with the inferred context set denoted by the yellow circles. The second row corresponds to the reconstructed images. The context sets inferred on the test images.}
\label{figure: }
\end{figure}
\subsection{FSQ-VAE}
\subsubsection{Training Dataset: t-ImageNet}
\begin{figure}[H]
\centering
\includegraphics[width=\linewidth, height=3.1cm]{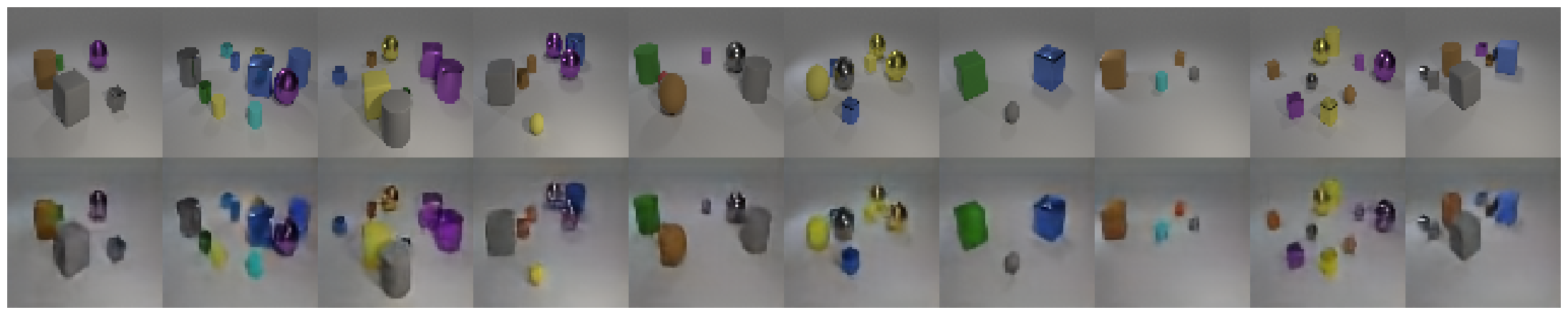}
\caption{The first row corresponds to the original image. The second row corresponds to the reconstructed images. Evaluated on the CLEVR test images.}
\label{figure: }
\end{figure}
\begin{figure}[H]
\centering
\includegraphics[width=\linewidth, height=3.1cm]{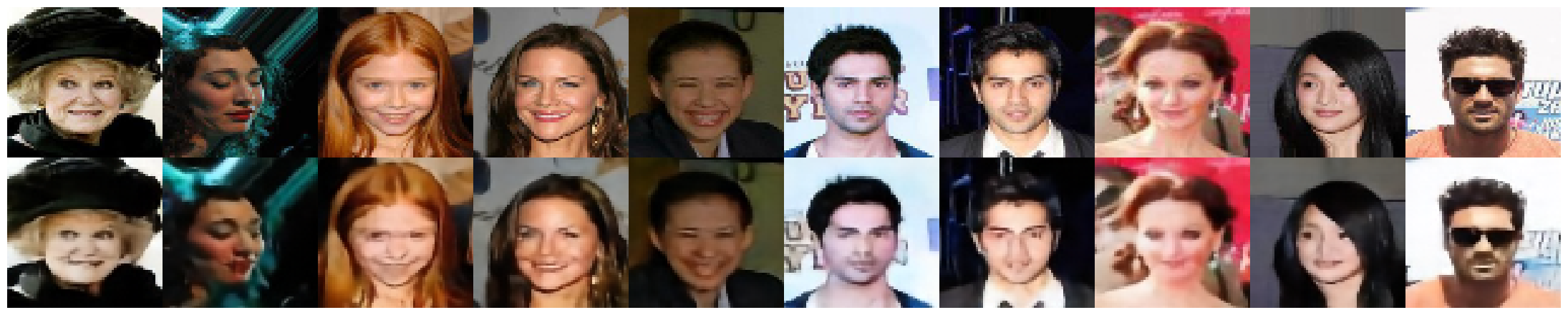}
\caption{The first row corresponds to the original image. The second row corresponds to the reconstructed images. Evaluated on the CelA test images.}
\label{figure: }
\end{figure}
\subsubsection{Training Dataset: CLEVR}
\begin{figure}[H]
\centering
\includegraphics[width=\linewidth, height=3.1cm]{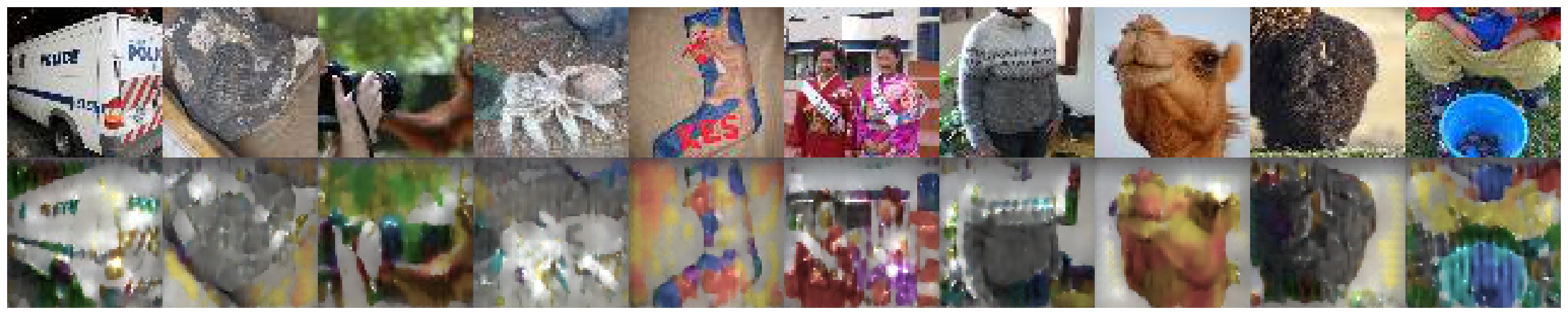}
\caption{The first row corresponds to the original image. The second row corresponds to the reconstructed images. Evaluated on the t-ImageNet test images.}
\label{figure: }
\end{figure}
\begin{figure}[H]
\centering
\includegraphics[width=\linewidth, height=3.1cm]{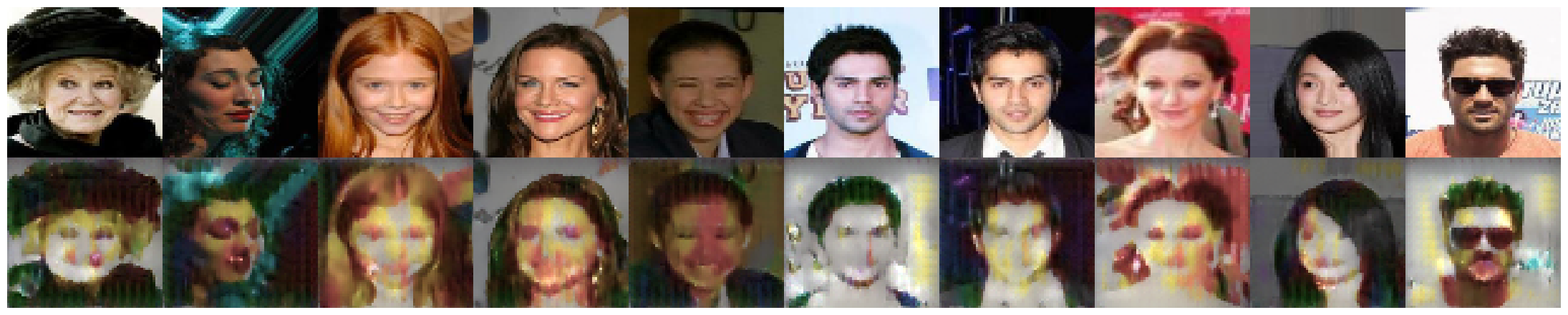}
\caption{The first row corresponds to the original image. The second row corresponds to the reconstructed images. Evaluated on the CelA test images.}
\label{figure: }
\end{figure}
\subsubsection{Training Dataset: CelebA}
\begin{figure}[H]
\centering
\includegraphics[width=\linewidth, height=3.1cm]{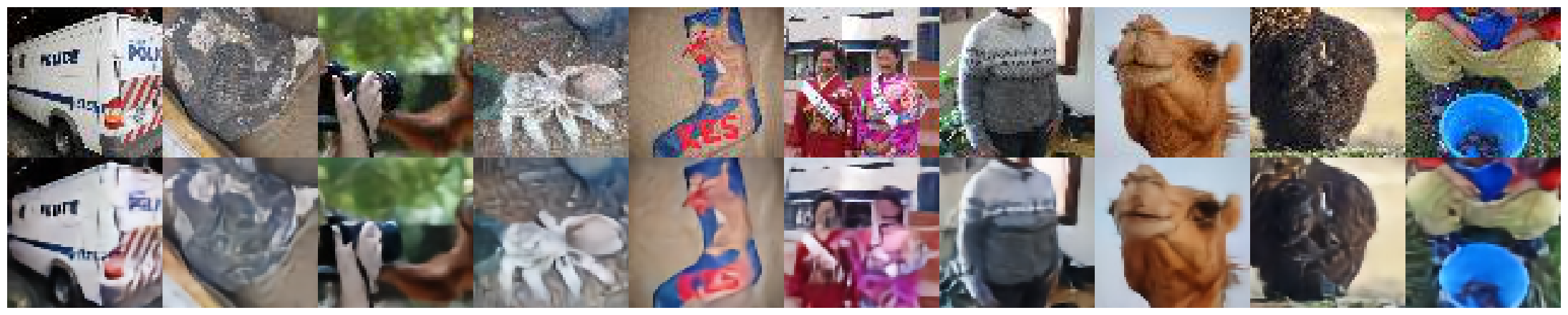}
\caption{The first row corresponds to the original image. The second row corresponds to the reconstructed images. Evaluated on the t-ImageNet test images. }
\label{figure: }
\end{figure}
\begin{figure}[H]
\centering
\includegraphics[width=\linewidth, height=3.1cm]{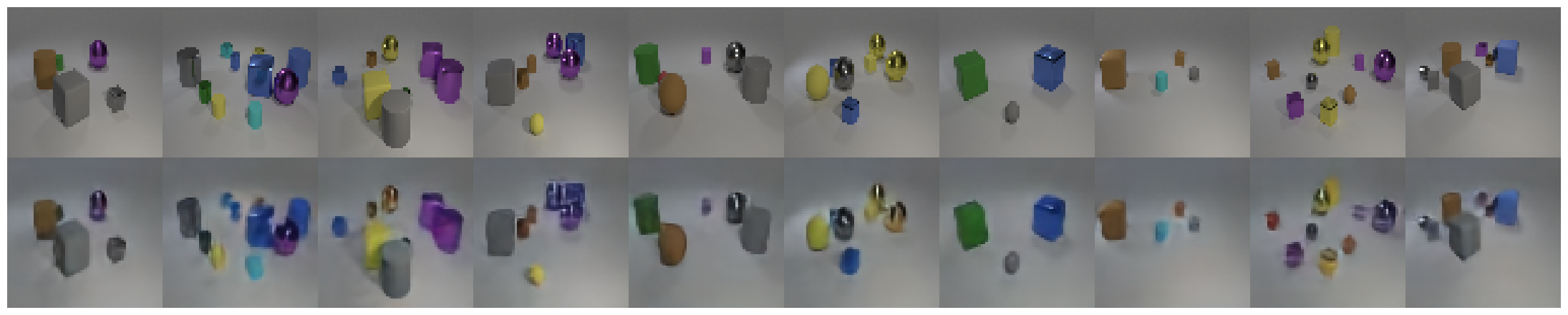}
\caption{The first row corresponds to the original image. The second row corresponds to the reconstructed images. Evaluated on the CLEVR test images.}
\label{figure: }
\end{figure}

\subsection{VQ-VAE}
\subsubsection{Training Dataset: t-ImageNet}
\begin{figure}[H]
\centering
\includegraphics[width=\linewidth, height=3.1cm]{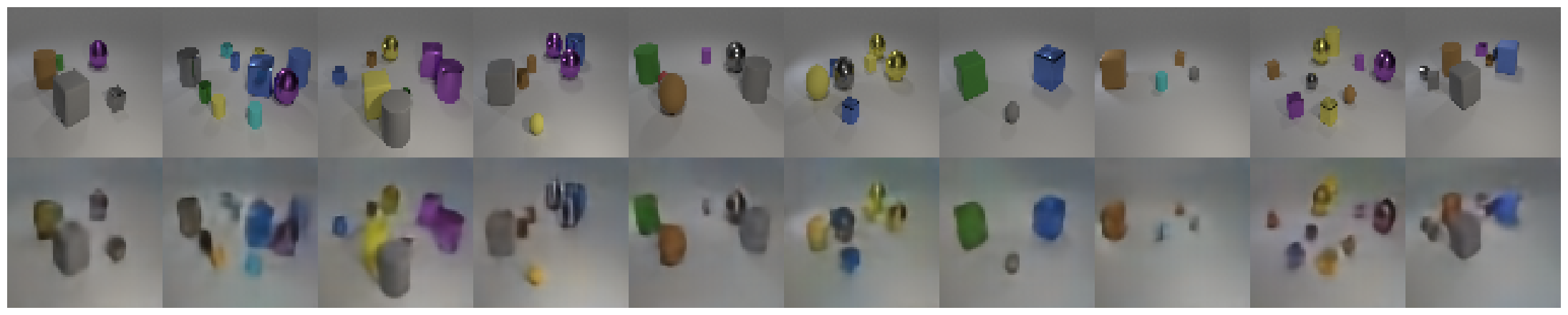}
\caption{The first row corresponds to the original image. The second row corresponds to the reconstructed images. Evaluated on the CLEVR test images.}
\label{figure: }
\end{figure}
\begin{figure}[H]
\centering
\includegraphics[width=\linewidth, height=3.1cm]{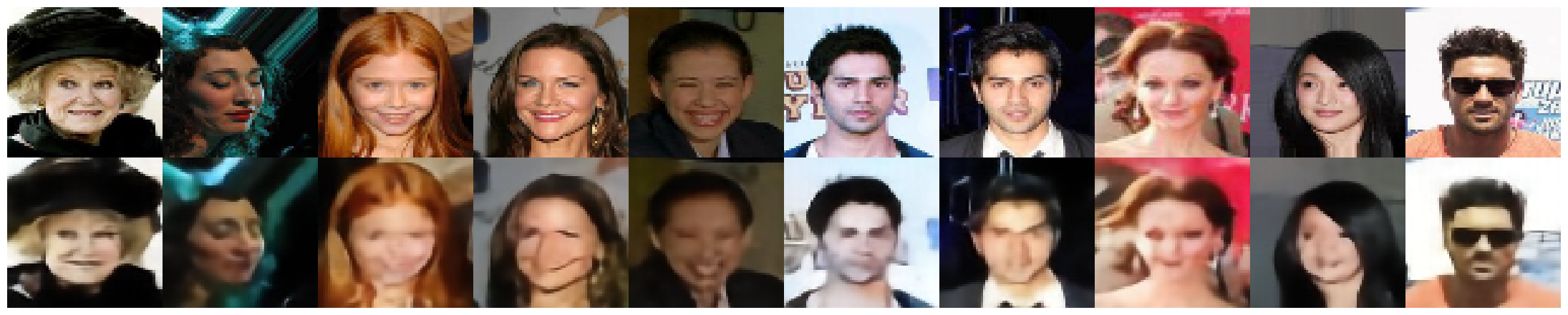}
\caption{The first row corresponds to the original image. The second row corresponds to the reconstructed images. Evaluated on the CelA test images.}
\label{figure: }
\end{figure}
\subsubsection{Training Dataset: CLEVR}
\begin{figure}[H]
\centering
\includegraphics[width=\linewidth, height=3.1cm]{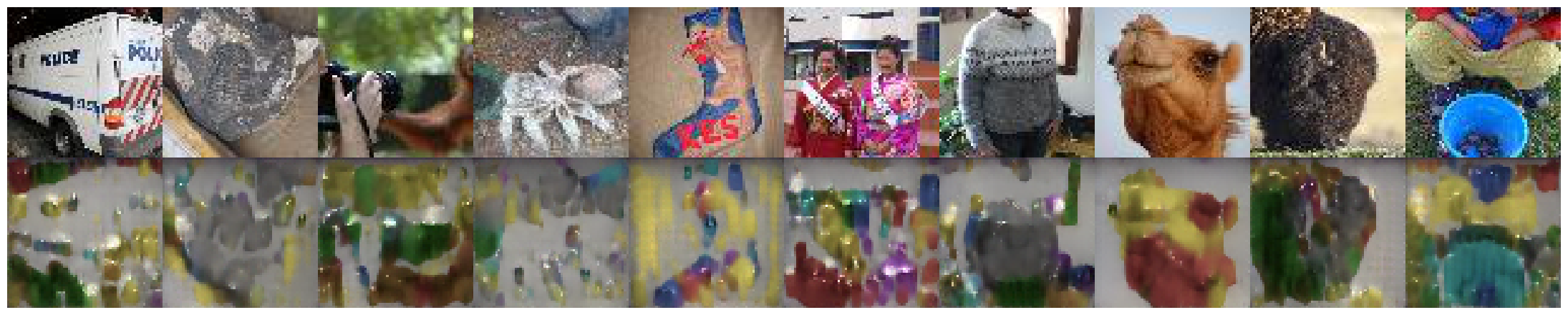}
\caption{The first row corresponds to the original image. The second row corresponds to the reconstructed images. Evaluated on the t-ImageNet test images.}
\label{figure: }
\end{figure}
\begin{figure}[H]
\centering
\includegraphics[width=\linewidth, height=3.1cm]{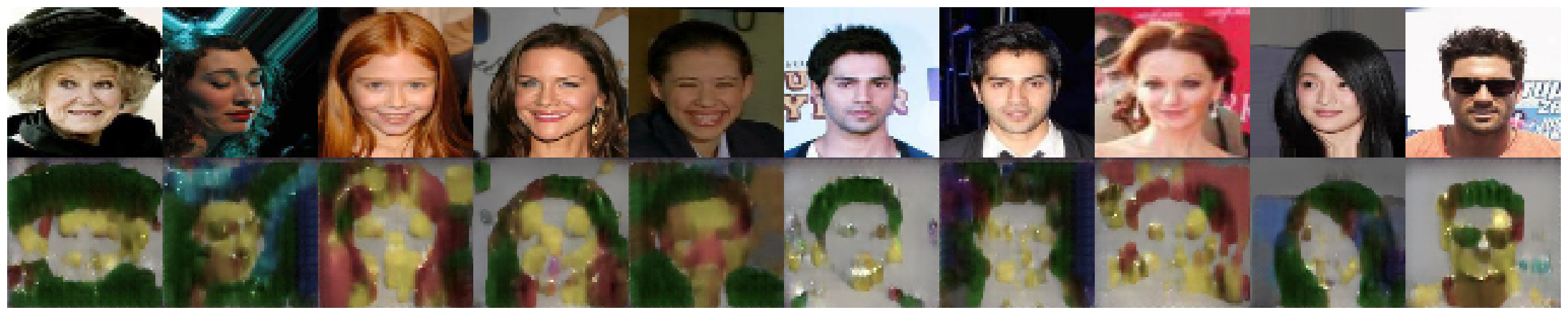}
\caption{The first row corresponds to the original image. The second row corresponds to the reconstructed images. Evaluated on the CelA test images.}
\label{figure: }
\end{figure}
\subsubsection{Training Dataset: CelebA}
\begin{figure}[H]
\centering
\includegraphics[width=\linewidth, height=3.1cm]{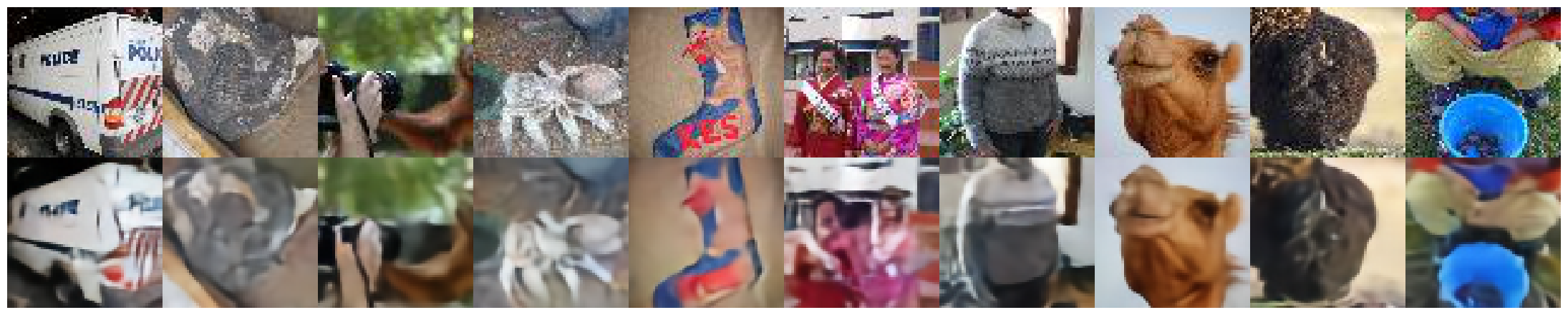}
\caption{The first row corresponds to the original image. The second row corresponds to the reconstructed images. Evaluated on the t-ImageNet test images.}
\label{figure: }
\end{figure}
\begin{figure}[H]
\centering
\includegraphics[width=\linewidth, height=3.1cm]{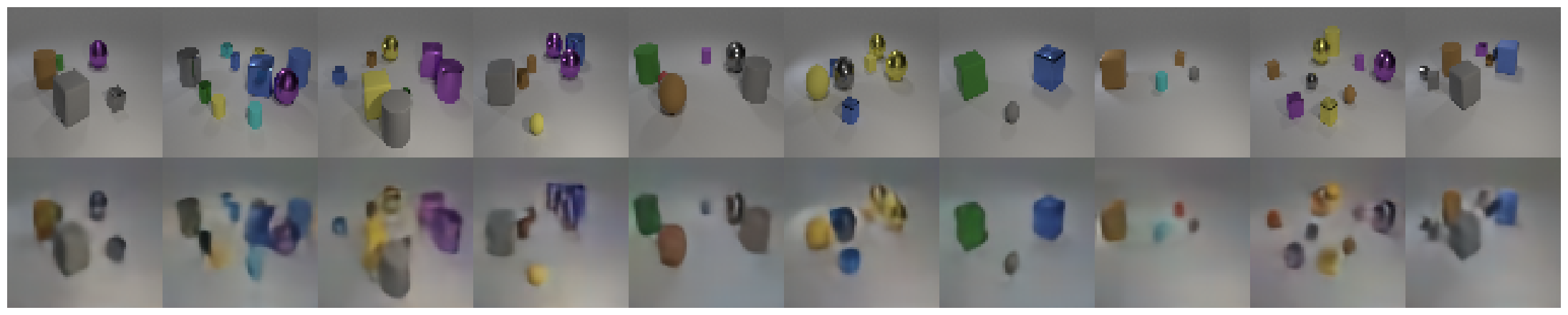}
\caption{The first row corresponds to the original image. The second row corresponds to the reconstructed images. Evaluated on the CLEVR test images.}
\label{figure: }
\end{figure}



\end{document}